\documentclass[runningheads]{llncs}
\usepackage{times}
\usepackage{epsfig}
\usepackage{graphicx}
\usepackage{amsmath}
\usepackage{amssymb}
\usepackage{svg}
\usepackage{makecell}
\usepackage{tikz}
\usetikzlibrary{spy}
\usepackage{booktabs}
\usepackage{fonttable}
\usepackage{tikz,pgfplots}
\usepackage{subcaption} 
\usepackage{multicol} 
\usepackage{soul}

\usepackage{graphicx}
\usepackage{multirow}
\usepackage{array}

\usepackage[normalem]{ulem}

\newcommand{\x}{\boldsymbol{\mathit{x}}}

\begin{document}
\pagestyle{headings}
\mainmatter

\def\ACCV22SubNumber{10}  

\title{Photorealistic Facial Wrinkles Removal} 
\titlerunning{ACCV-22 submission ID \ACCV22SubNumber}

\author{Marcelo Sanchez\inst{1,2} \and
Gil Triginer\inst{1} \and
Coloma Ballester\inst{2} \and
Lara Raad\inst{3} \and
Eduard Ramon\inst{1} \thanks{This work was done prior to joining Amazon.}}

\authorrunning{Marcelo Sanchez et al.}

\institute{Crisalix S.A\and
Universitat Pompeu Fabra \and
Université Gustave Eiffel}
{
\maketitle
\renewcommand\twocolumn[1][]{#1}%
\begin{center}
    \captionsetup{type=figure}
   \includegraphics[width=\textwidth]{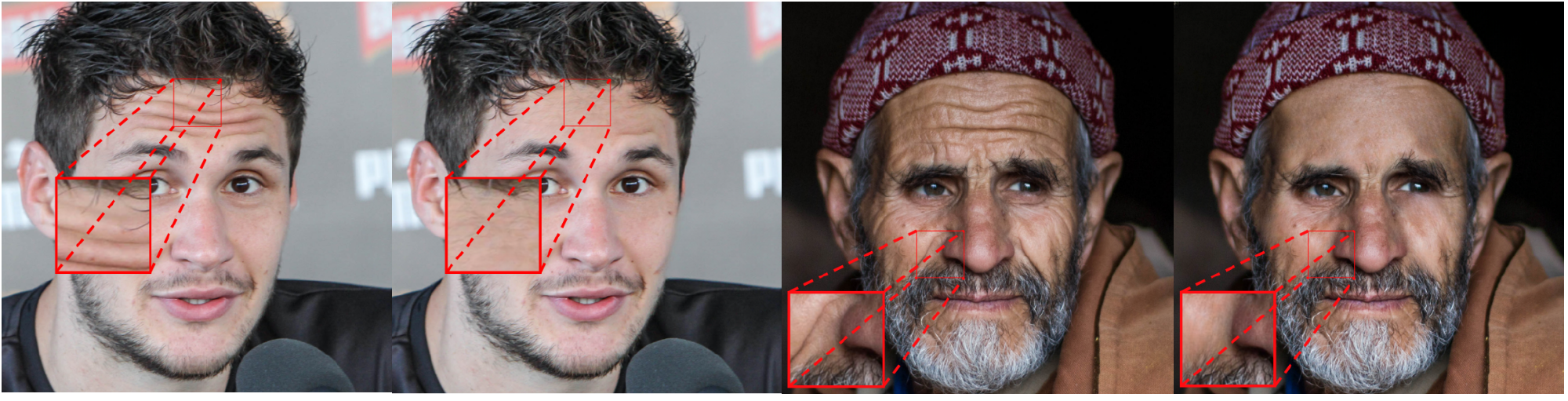}
    \captionof{figure}{Obtained results by our proposed wrinkle cleaning pipeline. Our method is able to obtain photorealistic results, learning to synthesize the skin distribution and obtain extremely realistic inpainting. Our model solves both wrinkle detection and wrinkle cleaning. For each of the two examples, the original image is shown on the left and the resulting image without wrinkles on the right, including a zoom-in of wrinkle regions.}
\end{center}
}
\begin{abstract}
Editing and retouching facial attributes is a complex task that usually requires human artists to obtain photo-realistic results. Its applications are numerous and can be found in several contexts such as cosmetics or digital media retouching, to name a few. Recently, advancements in conditional generative modeling have shown astonishing results at modifying facial attributes in a realistic manner. However, current methods are still prone to artifacts, and focus on modifying global attributes like age and gender, or local mid-sized attributes like glasses or moustaches. In this work, we revisit a two-stage approach for retouching facial wrinkles and obtain results with unprecedented realism. First, a state of the art wrinkle segmentation network is used to detect the wrinkles within the facial region. Then, an inpainting module is used to remove the detected wrinkles, filling them in with a texture that is statistically consistent with the surrounding skin. To achieve this, we introduce a novel loss term that reuses the wrinkle segmentation network to penalize those regions that still contain wrinkles after the inpainting. We evaluate our method qualitatively and quantitatively, showing state of the art results for the task of wrinkle removal. Moreover, we introduce the first high-resolution dataset, named \emph{FFHQ-Wrinkles}, to evaluate wrinkle detection methods.
\end{abstract}

\section{Introduction}
\label{sec:introduction}
\begin{figure*}[!ht]
\includegraphics[width=\textwidth, trim={0 0.3cm 0 0},clip]{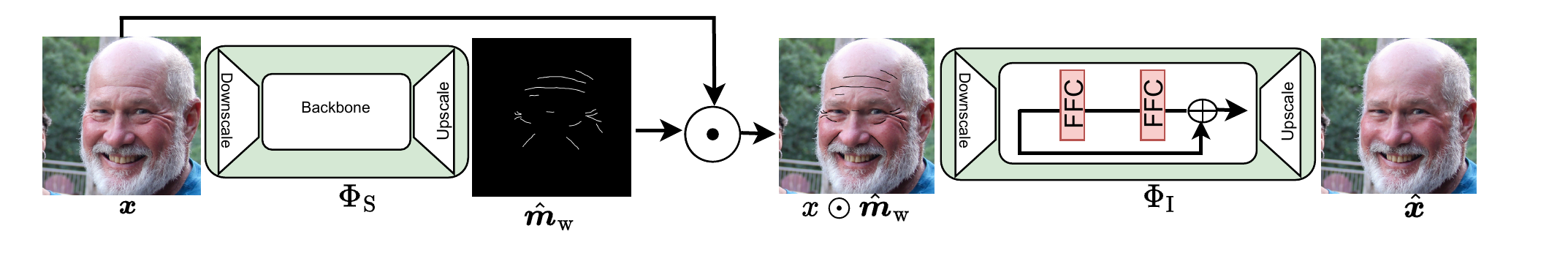}
\caption{Proposed pipeline for wrinkle removal. Input image \(\x\) is first forwarded through \(\boldsymbol{\phi_{S}}\) obtaining \(\hat{\boldsymbol{m}}_{w}\). This segmentation map \(\hat{\boldsymbol{m}}_{\rm w}\) is masked with \(\x\) and passed through the inpainting module \(\phi_{I}\), obtain the wrinkle-free color image \(\boldsymbol{\hat{x}}\).}
\label{fig:full-pipeline}
\end{figure*}

Facial image editing is a widely used practice, specially in cinema, television and professional photography. Many techniques require human artists to spend much time on manual editing using complex software solutions \cite{adobephotoshop} in order to obtain realistic modifications of the original image.
Removing skin imperfections, and specifically softening or removing wrinkles, is one of the most common tasks.

In order to ease this process, there have been general efforts on its automation \cite{visagelab}, and more concretely, in the automatic removal of facial wrinkles \cite{batool2014detection}. In \cite{batool2014detection} a two-stage approach is proposed in which the wrinkles are first detected using Gabor filters, and then removed using a texture synthesis method. While the  results from this approach are interpretable, the edited images lack photorealism and hardly maintain the statistics of the skin of the person.

Generative models based on deep learning, which learn the distribution of a particular category of data, have shown impressive results at generating novel faces and manipulating their content \cite{alaluf2021only,song2020face,lample2017fader,he2019attgan,ding2020injectiongan,karras2019style,wu2019relgan}. More concretely, Generative Adversarial Networks (GAN) \cite{goodfellow2014generative} can learn to generate highly realistic faces. As shown in \cite{abdal2020image2stylegan++}, it is possible to inpaint facial regions using GANs by optimizing the latent space to match a masked facial image, which in turn can be used to remove skin imperfections. However, this optimization process is slow.


Conditional generative models and in general image-to-image translation methods~\cite{isola2017image,zhu2017unpaired,ding2020injectiongan,liu2019stgan} provide a more flexible and fast framework than vanilla generative models, and allow conditioning the generation of new samples on input images and optionally on local semantic information. These methods have been successfully applied to the task of automatic professional face retouching~\cite{shafaei2021autoretouch} in combination with a large scale dataset with ground truth annotations. Recent work~\cite{suvorov2022resolution} introduces an image-to-image translation model for the task of image inpainting in which fast Fourier convolutions (FFC)~\cite{chi2020fast} are exploited, obtaining photorealistic inpainted images with a model learnt in an unsupervised fashion and that excels at modelling repetitive patterns, which is a desirable property for skin synthesis and wrinkle removal.



In this work, we propose a modern take on wrinkle removal by improving the standard pipeline proposed in \cite{batool2014detection} using data-driven techniques. First, we propose to replace the wrinkle detection stage using a state-of-the-art image segmentation neural network \cite{zhou2019unet++} in order to gain robustness and accuracy. In addition, we propose to replace the patch-based wrinkle cleaning block by an image inpainting network based on fast Fourier convolutions~\cite{suvorov2022resolution} to maintain the overall distribution of the person's skin in the modified area. 
In summary, our contributions are as follows:

\begin{itemize}
	\item A modern take on wrinkle removal using state-of-the-art segmentation and inpainting methods in order to obtain gains in accuracy and photorealism.
	\item A novel loss that leverages a segmentation network to supervise the inpainting training process, leading to improved results.
	\item{The first publicly available dataset for evaluating wrinkle segmentation methods.}
\end{itemize}

Our paper is organized as follows. In Section \ref{sec:relatec_work}, we review the state of the art for facial image editing and wrinkle removal. Next, in Section \ref{sec:method} we describe our two-stage solution for automatic wrinkle removal. In Section \ref{sec:experiments}, we provide quantitative and qualitative results, and show that our method obtains state-of-the-art results. Finally, in Section \ref{sec:conclusions} we end up with our conclusions.

 \section{Related work}
\label{sec:relatec_work}
\textbf{Wrinkle Removal.}
To the best of our knowledge, and also supported by a recent literature survey \cite{yap2021survey}, few works exist for wrinkle removal. An initial solution was proposed by Bastanfard et al.~\cite{bastanfard2004toward} using face anthropometrics  theory and image inpainting to erase the wrinkles.
The current state-of-the-art for wrinkles removal is~\cite{batool2014detection}, which follows a two-stage approach. The first stage detects wrinkles using an algorithm that combines Gabor features and texture orientation filters. Once the detection is obtained, they proposed an exemplar-based texture synthesis to inpaint the wrinkle regions. Beyond the novelty of the method, it only succeeds in half megapixels images due to memory constrains and the images obtained lack photorealism, introducing artifacts near the inpainting zone.
\\
\textbf{Image-to-Image Translation.}
Image-to-Image translation appeared with \cite{isola2017image} as a solution for solving the  mapping of an image from one domain to another. Isola et al. used a GAN-based scheme with a conditional setup. Several methods~\cite{isola2017image,zhu2017unpaired,ding2020injectiongan,liu2019stgan} are able to translate  domains that require to modify mid-size features such as glasses or mustache, while preserving the global structure of the face. Even though some of these methods need paired images in order to address the image translation problem \cite{isola2017image,shafaei2021autoretouch,NEURIPS2019_9015}, which is sometimes unfeasible in wrinkle removal, other pipelines can work with unpaired images \cite{zhu2017unpaired,park2020contrastive}. However, these methods tend to attend to global regions and are not able to modify small features such as wrinkles or other skin marks.
\\
\textbf{Image Inpainting.}
Image Inpainting aims to fill in missing or corrupted regions of an image so that the reconstructed image appears natural to the human eye. Accordingly, filled regions should contain meaningful details and preserve the original semantic structure. Image Inpainting is closely related to wrinkle cleaning in the sense that it allows modifying the desired (sometimes small-size) regions while preserving the uncorrupted, i.e. wrinkle-free, regions. Initially, deep learning based inpainting methods used vanilla convolutional neural networks (CNN). More complex methods regularize the structure of the filled regions with edges or segmentation maps~\cite{nazeri2019edgeconnect,liao2020guidance,yang2020learning,liao2021image}.

Many of these methods are based on CNNs with limited receptive fields, while it is well known that a large receptive field allows the network to better understand the global structure of an image. Recent work \cite{suvorov2022resolution} uses Fourier convolutions that introduce global context by optimizing on the frequency domain of the image. This allows the network to incorporate the texture patterns of the entire image and thus obtain a better inpainting result. Also this non-local receptive field allows the inpainting model to generalize to high resolution images.
\section{Method}
\label{sec:method}

Given a color image of a face  $\x \in \mathbb{R}^{H\times W\times 3}$, our goal is to remove all the wrinkles present in $\x$ while preserving photorealism. Ideally, we should only modify the facial areas with wrinkles and, at the same time, the modified regions should preserve the local statistics of the skin of the person.
We propose to solve both wrinkle removal and wrinkle cleaning via a two-stage model based on image segmentation and inpainting techniques.
First we estimate a segmentation map of all the wrinkles in the image using a state-of-the-art CNN segmentation model. Once this is obtained, we propose an inpainting model based on LAMA \cite{suvorov2022resolution} to fill in the wrinkles regions with photorealistic clear skin. We illustrate our pipeline in Fig.~\ref{fig:full-pipeline}.

\subsection{Wrinkle segmentation}
\label{section:detection}

In this first step, a segmentation model $\Phi_{\rm S}$ detects the wrinkles in $\x$, predicting a segmentation map $\hat{\boldsymbol{m}}_{\rm w} \in \mathbb{R}^{H\times W}$. The segmentation model $\Phi_{\rm S}$ is a state-of-the-art fully convolutional neural network with nested skip pathways, namely Unet++ ~\cite{zhou2019unet++}.
 
Wrinkle segmentation has an inherent problem of class imbalance, since usually the facial wrinkles occupy smaller regions on the image compared with the clear skin. 
We manage to tackle the class imbalance by using a region-based loss instead of a classical distribution-based one such as cross entropy. 
In particular, we choose to minimize the Dice loss, defined as
\begin{equation}
    \displaystyle\mathcal{L}_{\textrm{Dice}}(\boldsymbol{m},\boldsymbol{\hat{m}}) = \frac{2\sum_{i=1}^{H \times W}m_i \hat{m}_i}{\sum_{i=1}^{H \times W}m_i^2 + \sum_{i=1}^{H \times W }\hat{m}_i^2}
    \label{eq:dice}
\end{equation}
over a set of ground truth wrinkle annotations $\boldsymbol{m}_w$ and predicted segmentation maps $\hat{\boldsymbol{m}}_w = \Phi_{\rm S}(\x)$. For clarity, in~\eqref{eq:dice} we have omitted the subindices $w$  and the subindices $i$ indicate iteration over image pixels. 

\subsection{Wrinkle cleaning}
\label{section:cleaning}

Once the wrinkles have been detected using the module $\Phi_{\rm S}$, we aim at cleaning them using an inpainting neural network $\Phi_{\rm I}$. The resulting inpainted image $\hat{\x}$ must be photorealistic and the generated skin must be statistically similar to the skin of the input image subject.


The regions to be inpainted are defined on a binary mask $\boldsymbol{m}_{\rm I} \in \mathbb{R}^{H\times W}$. We mask the input image $\x \in \mathbb{R}^{H\times W\times 3}$ and obtain the masked image $\x \odot \boldsymbol{m}_{\rm I}$. In order to add redundancy, we follow the approach from \cite{suvorov2022resolution} and concatenate $\boldsymbol{m}_{\rm I}$ to the masked image $\x \odot \boldsymbol{m}_{\rm I}$, obtaining a 4-channel input $\boldsymbol{x'} = \texttt{stack}(\x \odot \boldsymbol{m}_{\rm I},\boldsymbol{m}_{\rm I})$. Therefore, our wrinkle-free image is $\boldsymbol{\hat{x}} =  \Phi_{\rm I}(\boldsymbol{x'})$ where $\boldsymbol{\hat{x}} \in \mathbb{R}^{H\times W\times 3}$ and \(\Phi_{\rm I}\) corresponds to our inpainting module. In the following, we describe the training of $\Phi_{\rm I}$, which is illustrated in Fig.~\pageref{fig:train-pipeline}.

\begin{figure*}[ht]
\includegraphics[width=\textwidth, trim={0 0.3cm 0 0},clip]{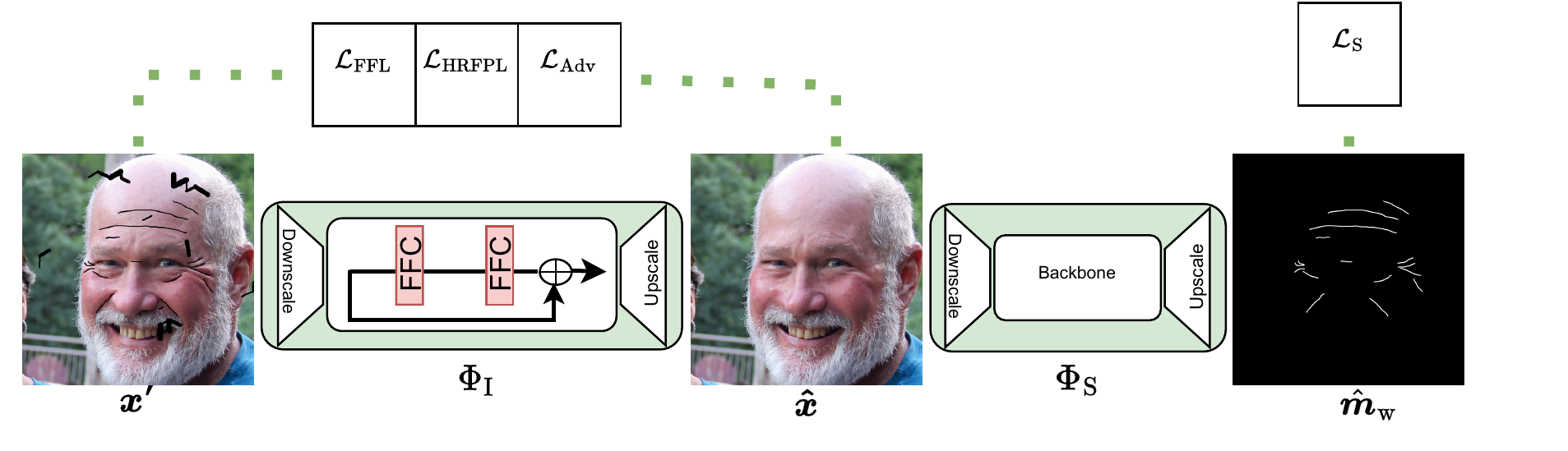}
\label{fig:train-pipeline}
\caption{Scheme of the proposed inpainting training pipeline. Image $\boldsymbol{x'}$ is forwarded through the inpainting module. The resulting inpainted image is used in multi-component loss.}
\end{figure*}


\textbf{Mask generation during training} In order to  train $\Phi_{\rm I}$, we generate pairs of images and masks. The inpainting masks $\boldsymbol{m}_{\rm I}$ are generated as the pixelwise logical union of two binary masks: the first marking pixels with wrinkles, $\boldsymbol{m}_{\rm w}$, and the second, $\boldsymbol{m}_{\rm g}$, marking wrinkle-shaped regions randomly generated using polygon chains \cite{suvorov2022resolution}. The latter is added in order to provide more diverse training data and wrinkle-free skin regions. The mask generation policy plays a big role in inpainting pipelines~\cite{suvorov2022resolution,wang2018high} and influences to a high extent the final performance of the model. The training mask policy should generate masks as similar as possible to the the inpainting regions at test time. A huge difference between training and testing masks can lead to a poor inpainting generalization.


We leverage the LAMA architecture~\cite{suvorov2022resolution} in our inpainting module. The underlying components of this network are fast Fourier convolutions (FFC)~\cite{chi2020fast}. While CNNs have a limited local receptive field, and use kernels that operate at a fixed spatial scale, FFCs capture global information of the input image and reuse the same kernels at different frequencies (scales). These are useful properties in wrinkle inpainting, due to the repetitive and multi-scale nature of skin texture.


\textbf{Training loss} The inpainting problem is ill-posed, as multiple image restorations can be appealing to the human eye. Because of this, training inpainting models is normally a challenging problem~\cite{suthar2014survey}, sometimes needing multiple losses in order to obtain good results. Simple supervised losses such as the squared Euclidean norm $L_2$ force the generator to inpaint the corrupted regions precisely. However, in some cases the visible parts do not provide enough information to recover the corrupted part successfully, leading to blurry results. More complex losses like the perceptual loss \cite{johnson2016perceptual} evaluate distance in the feature space of a pre-trained network \(\varphi(\cdot)\), encouraging reconstructions less equal to the ground truth in favor of global perceptual structure. In order to obtain photorealistic results, we also make use of multiple loss terms, which we define next.

First, we use a perceptual loss based on a high receptive field network $\varphi_{\rm{HRF}}(\cdot)$ trained on an image segmentation task \cite{suvorov2022resolution}. Namely, we use: 

\begin{equation}\label{eq:eq2}
    \mathcal{L}_{\textrm{HRFPL}}(\boldsymbol{x},\boldsymbol{\hat{x}}) = {\Upsilon}([\varphi_{\rm{HRF}}(\boldsymbol{x}) - \varphi_{\textrm{HRF}}(\hat{\boldsymbol{x}})]^2
    \odot(1 -  \texttt{resize}(\boldsymbol{m}_{w}))),
\end{equation}
where $[\cdot - \cdot]^2$ is the squared element-wise difference operator and $\Upsilon$ is the two stage mean operation, inter-layer mean of intra-layer means. Recall that the use of \texttt{resize} operator is needed in order to match the spatial dimensions of the mask with the deep features from $\varphi_{\rm{HRF}}$. All channel features are equally weighted to 1. It is important to note that since we can not acquire ground truth
images with the wrinkles cleaned, the supervised loss $\mathcal{L}_{\textrm{HRFPL}}$ is not computed on the wrinkle pixels defined by the non-zero pixel-values in $\boldsymbol{m}_{\rm{w}}$, in order to prevent the model from learning to generate wrinkles. Besides, following the well-known adversarial strategy, we use a discriminator \(D\) to ensure coherent and structured images generated by our inpainting module. We use the approach proposed in~\cite{wang2018high} using a patch-GAN discriminating at a patch-level. Accordingly, the corresponding loss terms are defined as:
    \begin{equation}
    \begin{aligned}
    \mathcal{L_D}
    &= -\mathbb{E}_{\boldsymbol{x}\sim {\mathbb{P}}}[\log D(\boldsymbol{x})] - \mathbb{E}_{\boldsymbol{\hat{x}}\sim {\mathbb{P}}_{\phi_{I}}}[\log D ( \boldsymbol{\hat{x}}) \odot \boldsymbol{m}]\\ 
    &\qquad - \mathbb{E}_{\boldsymbol{\hat{x}}\sim {\mathbb{P}}_{\phi_{I}}}[\log (1 - D(\boldsymbol{\hat{x}})) \odot (1 - \boldsymbol{m})],\\
   \mathcal{L_{\phi_{I}}} 
    &= - \mathbb{E}_{\boldsymbol{\hat{x}}\sim {\mathbb{P}}_{\phi_{I}}}[\log D ( \boldsymbol{\hat{x}})],\\
    \mathcal{L_\textrm{Adv}} &=  \mathcal{L_D} + \mathcal{L_{\phi_{I}}},
    \end{aligned}
    \end{equation}
where $\boldsymbol{x}$ denotes a sample from the dataset distribution  ${\mathbb{P}}$, and $\hat{\boldsymbol{x}}$ a sample from the inpainted images distribution ${\mathbb{P}_{\phi_{I}}}$.  

The losses presented so far are optimized in the spatial domain. In order to enforce similarity in the spectral domain and encourage high frequencies, we add the Focal Frequency Loss (FFL) defined in~\cite{jiang2021focal}. This frequency domain loss enforces $\Phi_{I}$ to recover the Fourier modulus and phase of the input image. Concurrently, the authors of~\cite{lu2022glama} have used the same strategy. The FFL is computed as:
\begin{equation}
    \mathcal{L_{\textrm{FFL}}} = \frac{1}{MN} \sum_{u=0}^{M-1}\sum_{v=0}^{N-1} w(u,v)|F_{\boldsymbol{\mathit{x}}}(u,v) - F_{\boldsymbol{\mathit{\hat{x}}}}(u,v)|^2,
\end{equation}
where $F_{\boldsymbol{\mathit{x}}}$ denotes the 2D discrete Fourier transform operator of the input image $\boldsymbol{\mathit{x}}$ and $F_{\boldsymbol{\mathit{\hat{x}}}}$ denotes the 2D discrete Fourier transform of the inpainted image $\boldsymbol{\mathit{\hat{x}}}$. The matrix $w(u,v)$ correspond to the weight of the spatial frequency at coordinates $(u,v)$ and are defined as $w(u,v) = |F_{\boldsymbol{\mathit{x}}}(u,v) - F_{\boldsymbol{\mathit{\hat{x}}}}(u,v)|$. The gradient through the matrix $w$ is locked, only serving as a weight for each frequency.
 
Up to this point, none of the presented losses are specific to the problem of wrinkle removal. We propose a novel \textit{ wrinkle loss} $\mathcal{L_S}$ for inpainting that creates a connection between the segmentation and the inpainting module. Inspired by the fact that our final goal is to generate images without wrinkles, we make use of our wrinkle segmentation module $\boldsymbol{\Phi_{\rm S}}$ to detect wrinkles in the inpainted image $\boldsymbol{\mathit{\hat{x}}}$ and to generate a supervision signal that discourages the presence of wrinkles in the inpainted image. More concretely, we first use the previously trained segmentation module $\boldsymbol{\Phi_{\rm S}}$ to segment $\boldsymbol{\mathit{\hat{x}}}$ and obtain a wrinkle map. Then, we enforce that the generated wrinkle map does not contain any wrinkle pixels. The loss term $\mathcal{L_S}$ is defined as follows:

\begin{equation}
    \mathcal{L_\textrm{S}}(x) = \Psi({\Phi_{\rm S}(\Phi_{\rm I}}(\boldsymbol{x}))),
\end{equation}
where $\Psi$ is the spatial mean operator.

In addition to the introduced losses, we also make use of the gradient penalty loss \cite{ross2018improving} defined as $\mathcal{R_{\textrm{1}}} = \mathbb{E}_{\boldsymbol{x}\sim {\mathbb{P}}}||\nabla D(\boldsymbol{x})||^2$, and the feature matching loss  $\mathcal{L_{\textrm{DiscPL}}}$ which is similar to $\mathcal{L}_{\textrm{HRFPL}}$ but the features are obtained from the discriminator \(D\), which is known to stabilize training \cite{wang2018high}. The final loss function for our wrinkle inpainting model can be written as:
\begin{equation}
    \begin{aligned}
    \mathcal{L}
    & = \lambda_0 \mathcal{L_{\textrm{Adv}}} + \lambda_1 \mathcal{L_{\textrm{HRFPL}}} + \lambda_2 \mathcal{L_{\textrm{DiscPL}}} +
     \lambda_3  \mathcal{R_{\textrm{1}}} + \lambda_4 \mathcal{L_{\textrm{FFL}}} +\lambda_5 \mathcal{L_\textrm{S}},
    \end{aligned}
\end{equation}
where $\lambda_{i} >0, ~i\in\{0,\dots,5\}$ 
are fixed hyperparameters (detailed in Section~\ref{sec:experiments}). 

Even though our pipeline solves wrinkle segmentation and wrinkle cleaning, both solutions are not learnt simultaneously. The two problems require different learning strategies and have different time of convergence, applying simultaneous learning would damage the global accuracy of the system. 


\subsection{Inference} The segmentation and inpainting module are used sequentially to obtain the output image $\boldsymbol{\hat{x}}$. First, we forward the image through the segmentation model $\Phi_{\rm S}$ to obtain the wrinkle map. With the segmentation map we then mask our input image and generate the inpainted image through $\Phi_{\rm I}$. A schematic representation of the inference process is displayed in Fig.~\pageref{fig:train-pipeline}.

\begin{figure*}[!ht]
\setlength{\tabcolsep}{1pt}
\begin{tabular}{cccccc}
\includegraphics[height=0.25\linewidth,width=0.25\linewidth]{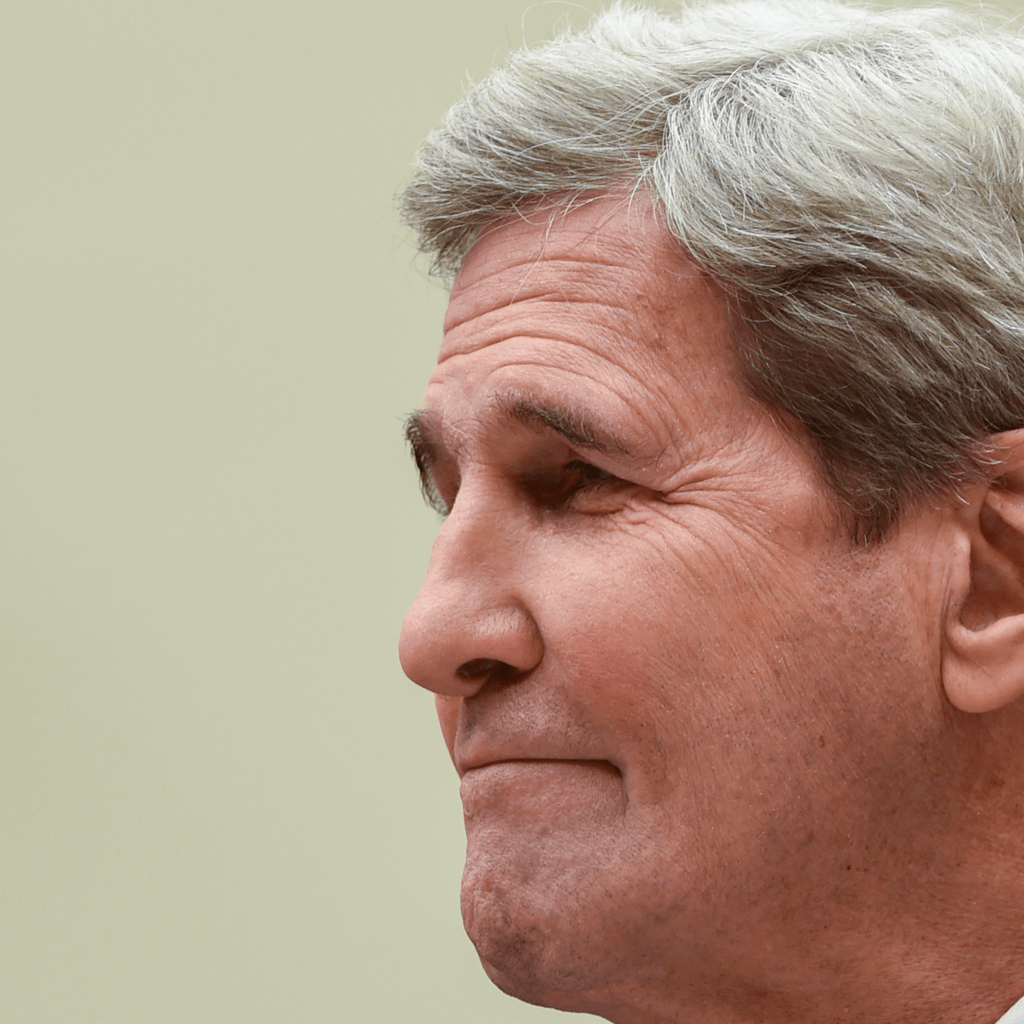} &
\includegraphics[height=0.25\linewidth,width=0.25\linewidth]{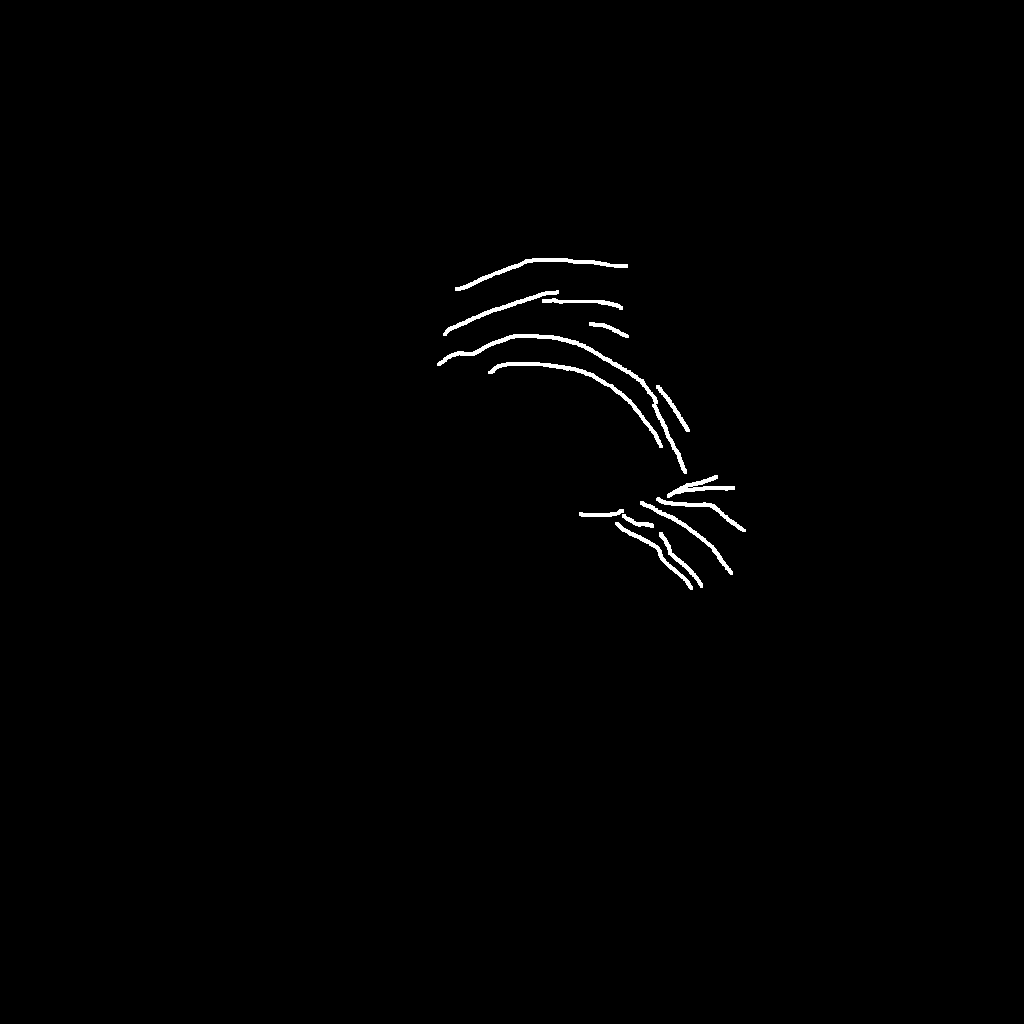} &
\includegraphics[height=0.25\linewidth,width=0.25\linewidth]{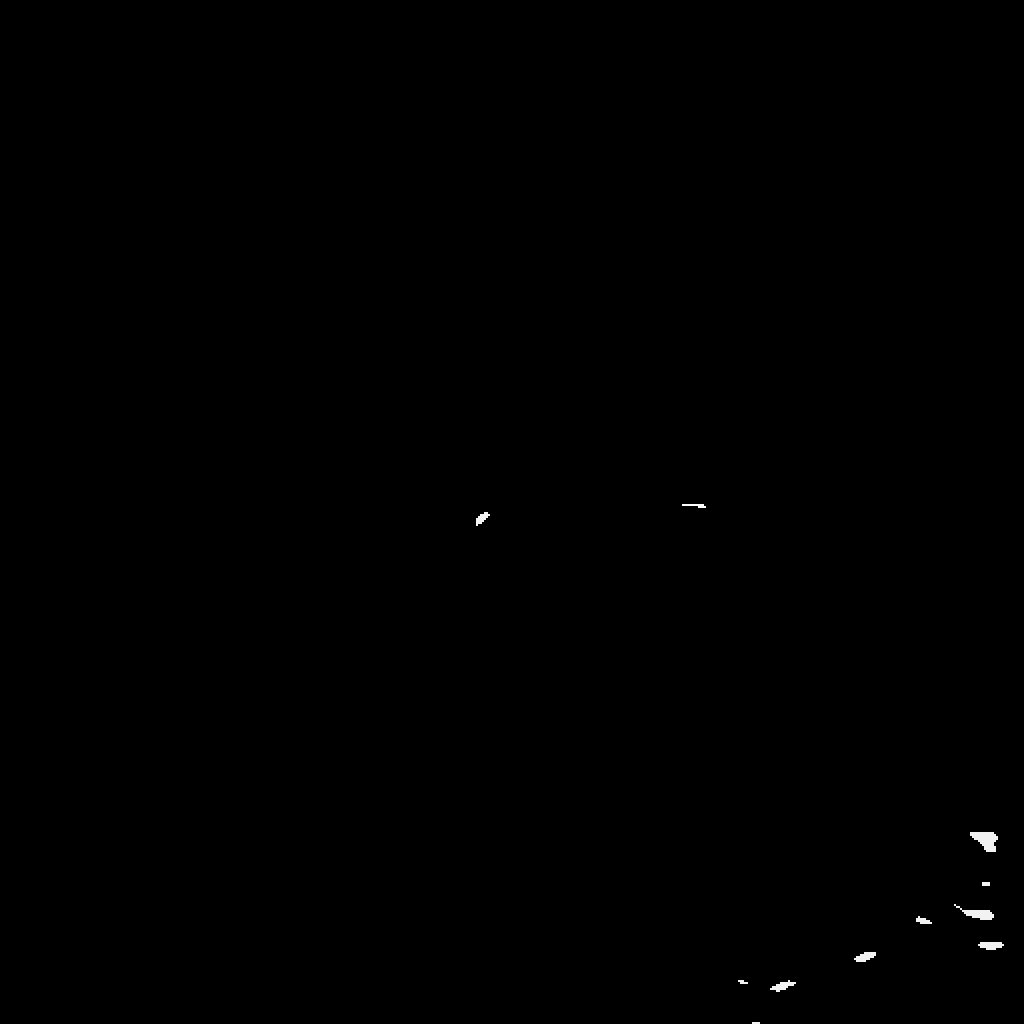} &
\includegraphics[height=0.25\linewidth,width=0.25\linewidth]{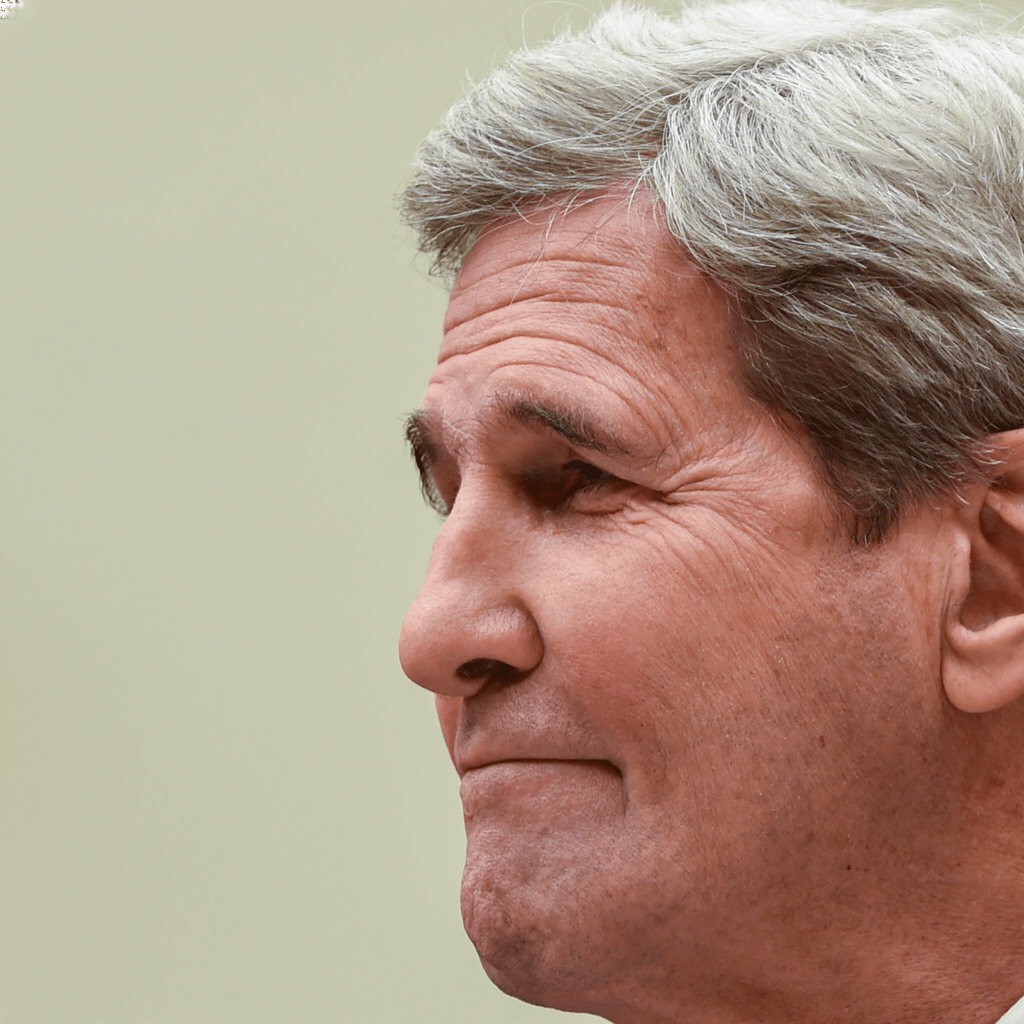} & \\
\includegraphics[height=0.25\linewidth,width=0.25\linewidth]{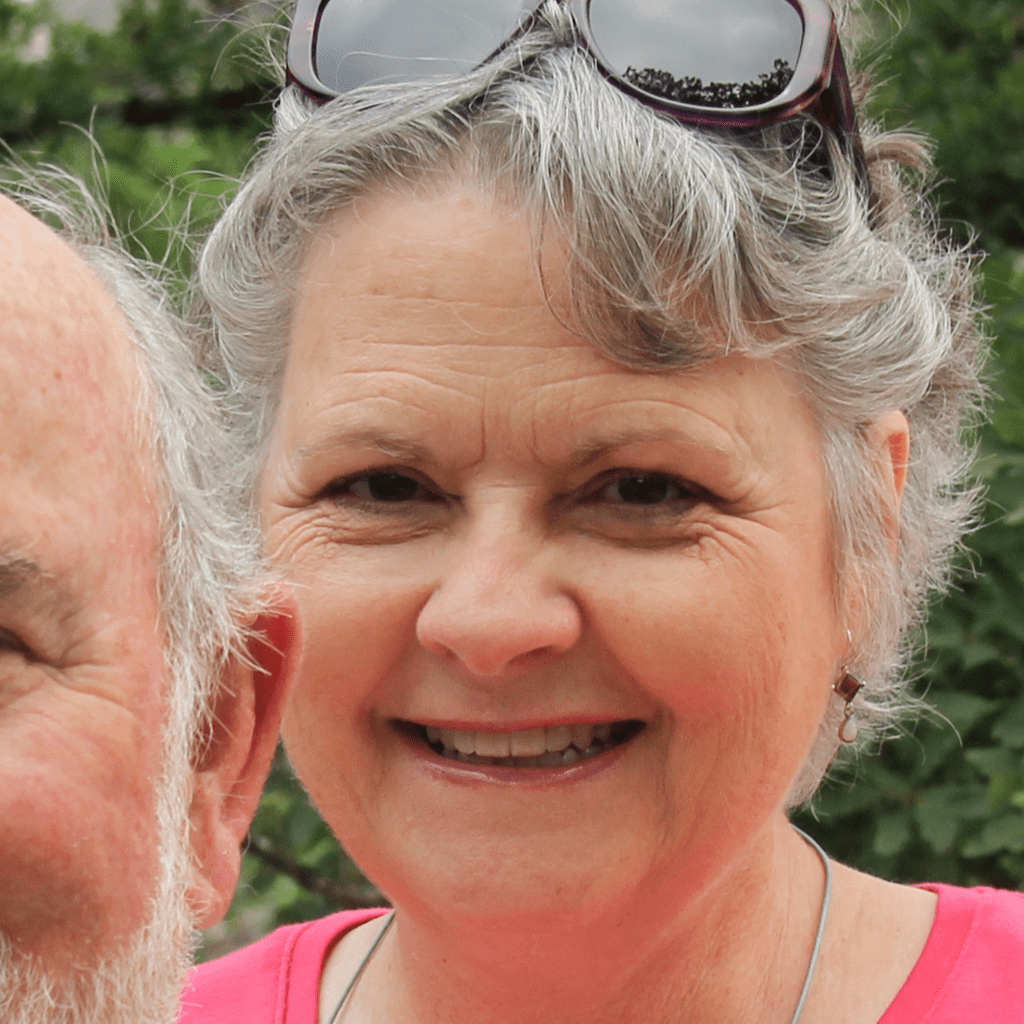} &
\includegraphics[height=0.25\linewidth,width=0.25\linewidth]{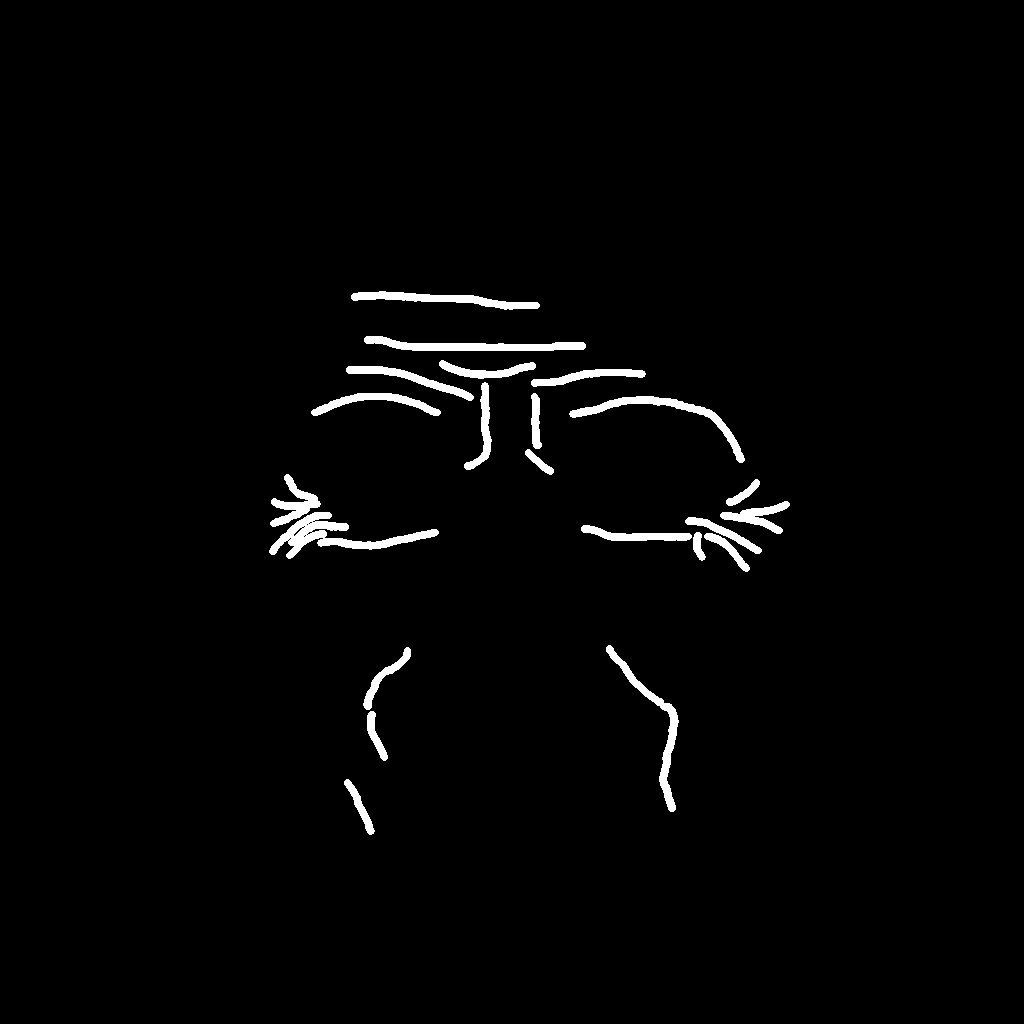} &
\includegraphics[height=0.25\linewidth,width=0.25\linewidth]{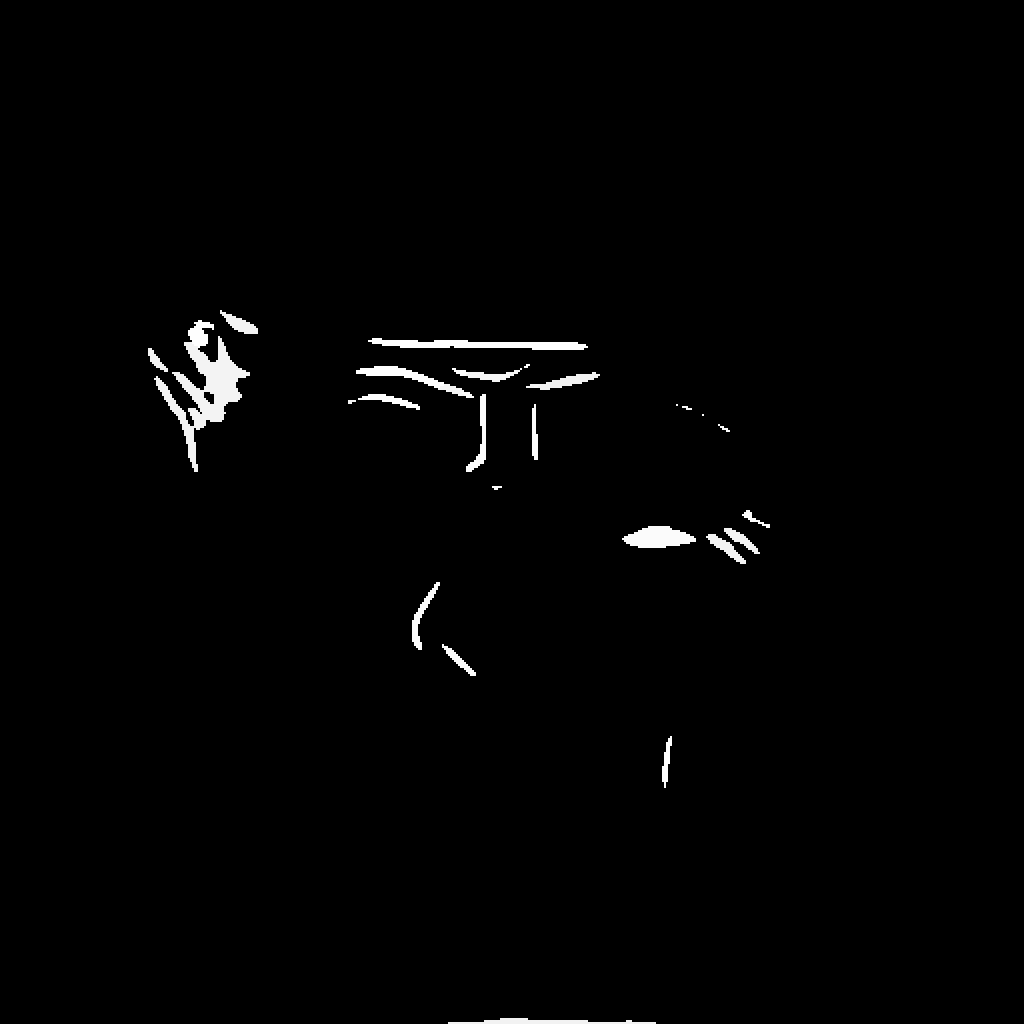} &
\includegraphics[height=0.25\linewidth,width=0.25\linewidth]{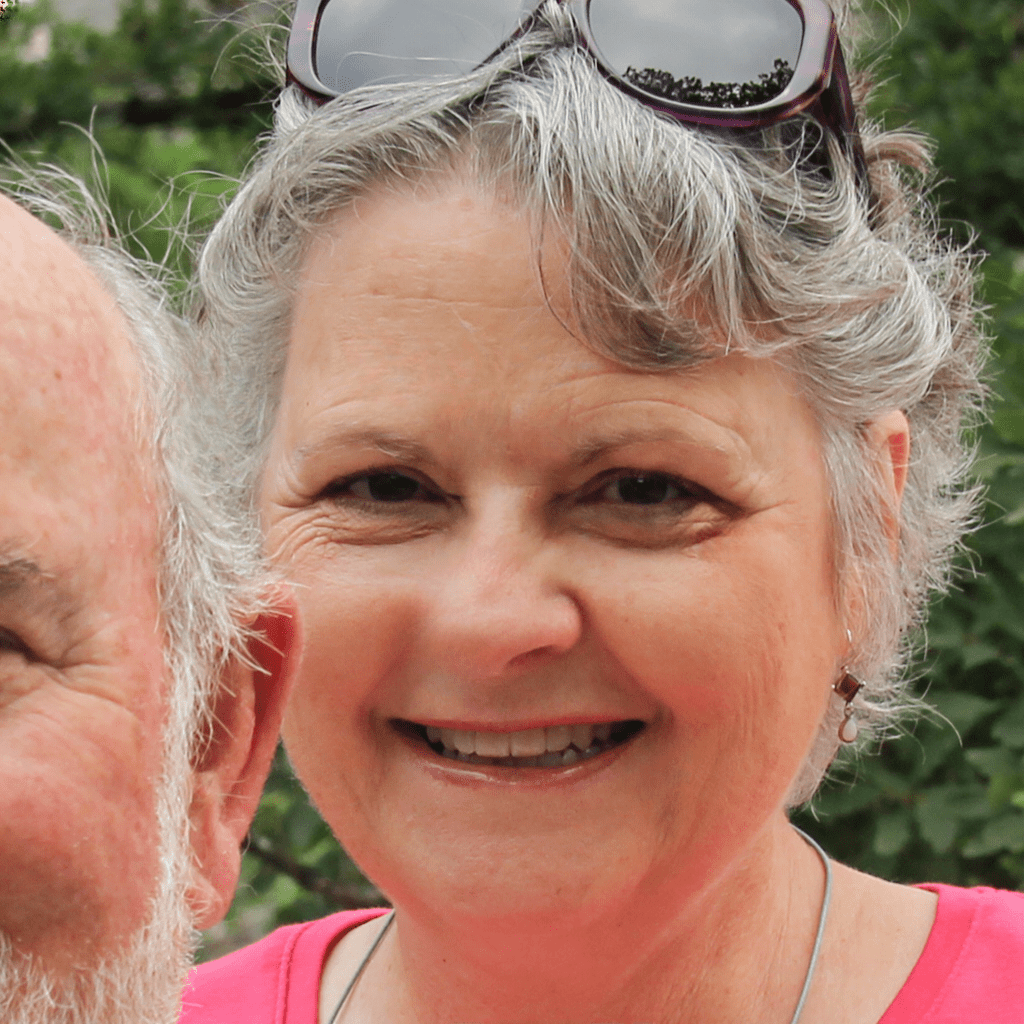} & \\

{{Image}} & {\makecell{ground truth \\ segmentation}} & {\makecell{predicted \\ segmentation}} & {{Result}}
\end{tabular}
\caption{Qualitative results of our wrinkle cleaning pipeline. This examples display the limitations of our method and the importance of a correct segmentation in the wrinkle cleaning pipeline.}
\label{fig:limitations}
\end{figure*}

However, the inpainting results are always limited by the performance of the segmentation module. The inpainting is always conditioned by the segmentation map $\hat{\boldsymbol{m}}_{\rm{w}}$, wrinkles that are not detected  by $\Phi_{\rm S}$ will remain in $\boldsymbol{\hat{x}}$. This problem can be observed in Fig.~\ref{fig:limitations}
\section{Experimental results}
\label{sec:experiments}

In this section we provide quantitative and qualitative results showing the effectiveness of our method. Next, we introduce the datasets and the metrics that will be used along the section.

\subsection{Dataset and metrics}

\begin{figure}[t!]
    \centering
    \begingroup
    \setlength{\tabcolsep}{1pt} 
    \begin{tabular}{cccccc}
    \hspace*{-0.5in}
    \includegraphics[height=0.2\linewidth]{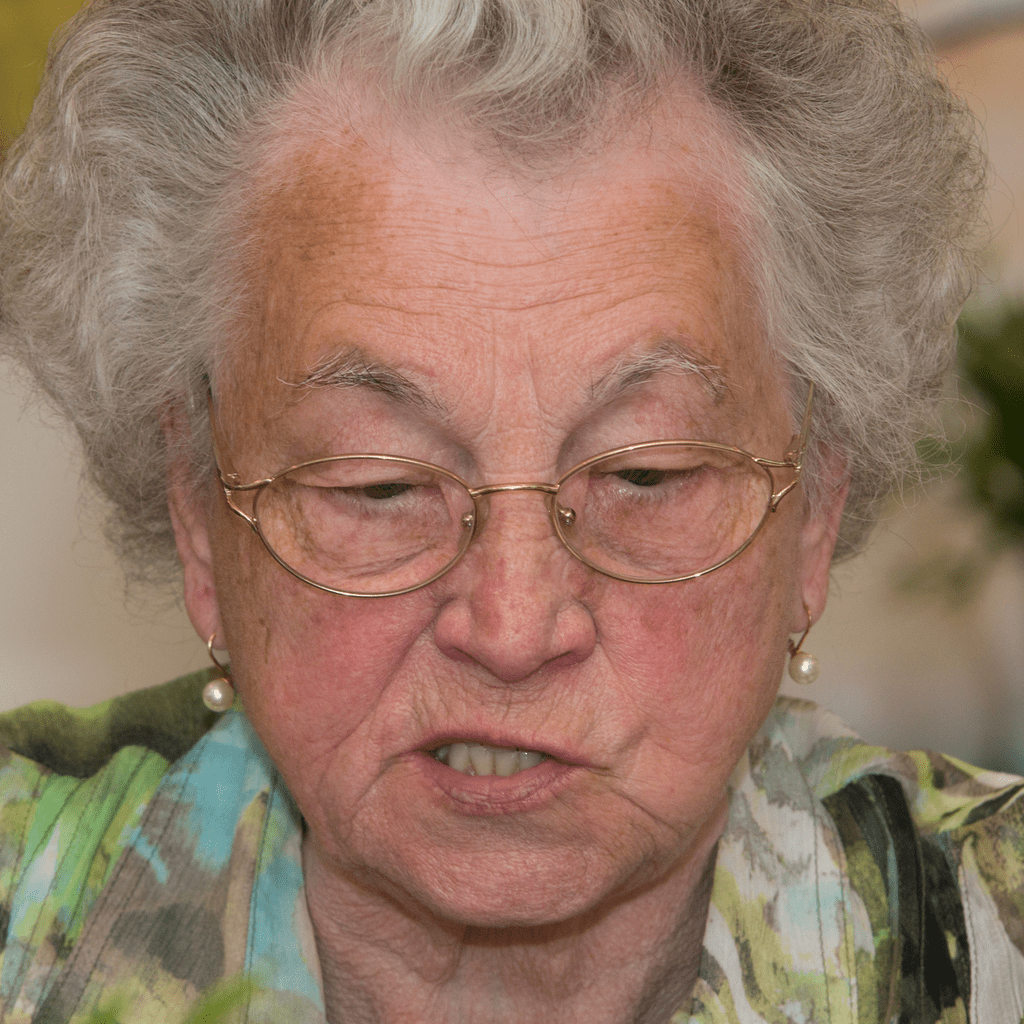} &
     \includegraphics[height=0.2\linewidth]{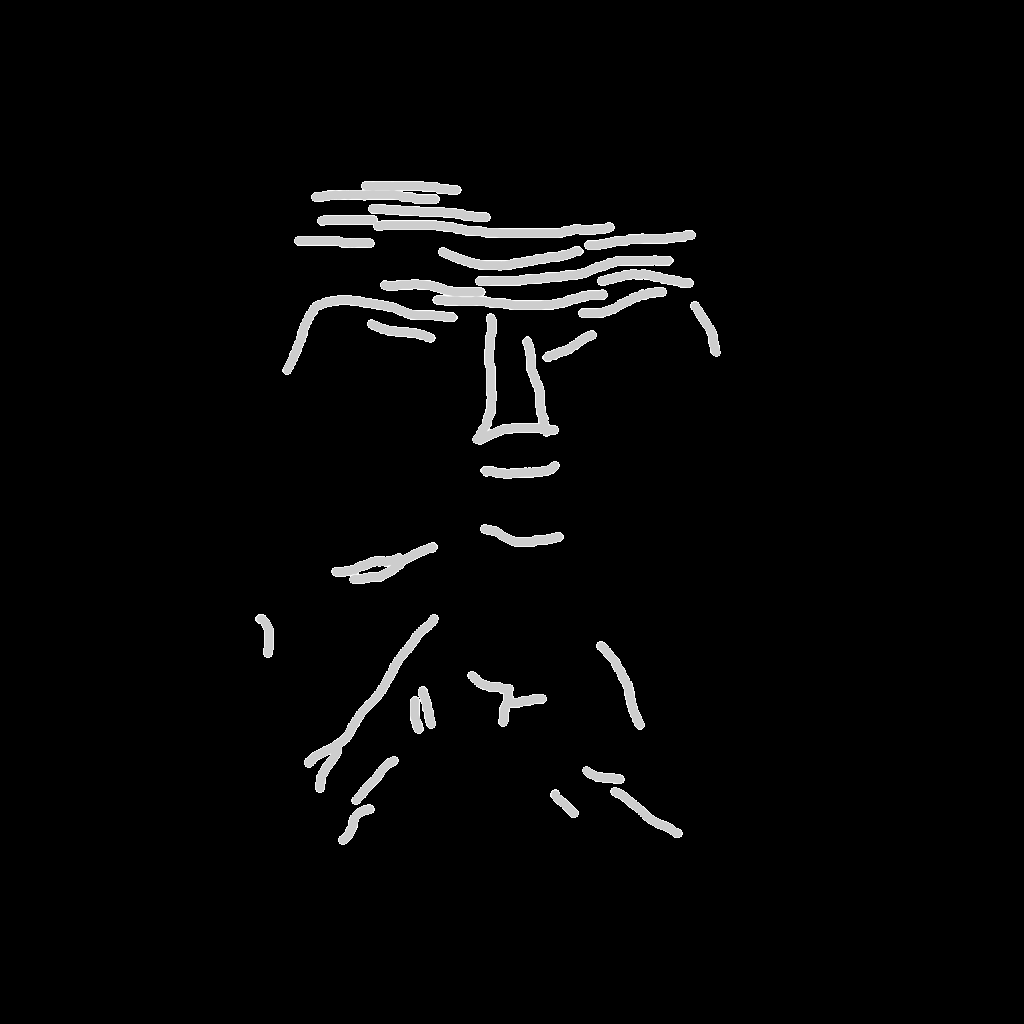} &
     \includegraphics[height=0.2\linewidth]{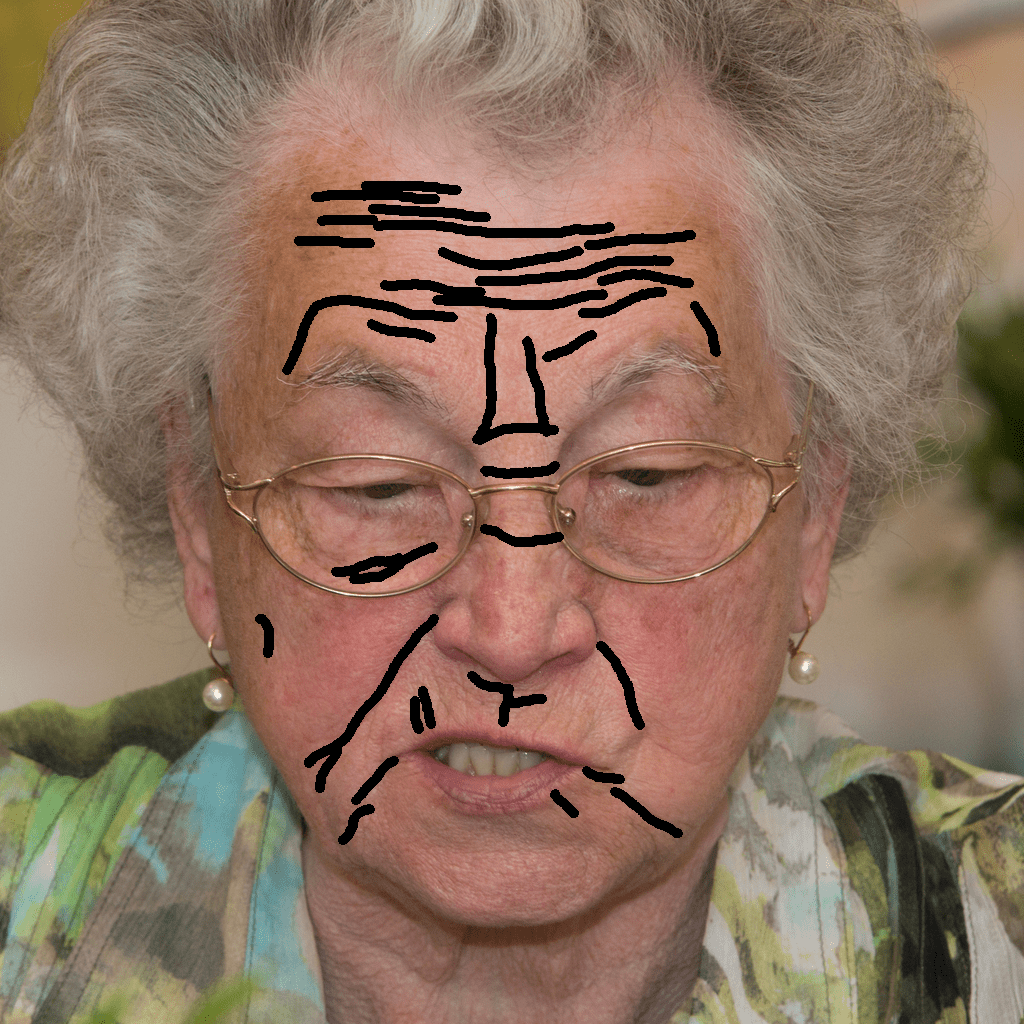}

     \includegraphics[height=0.2\linewidth]{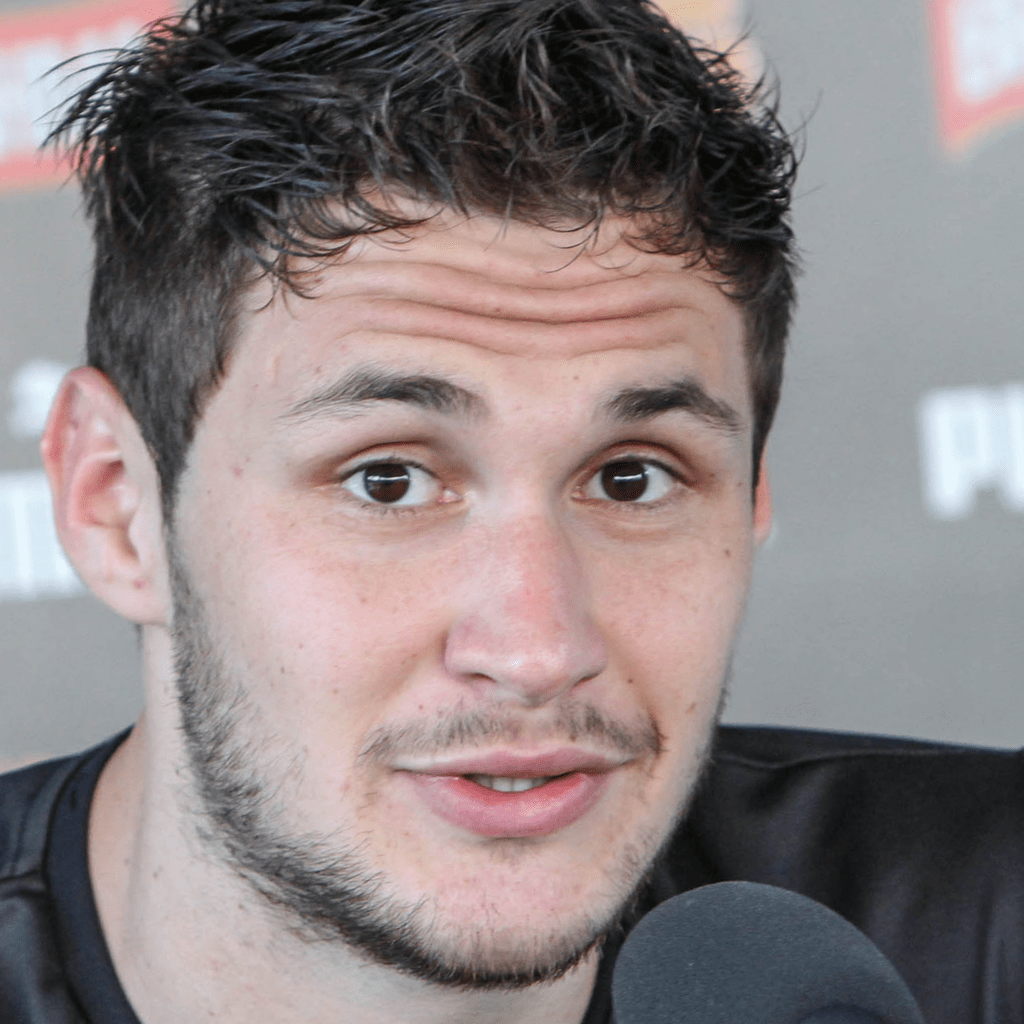} &
     \includegraphics[height=0.2\linewidth]{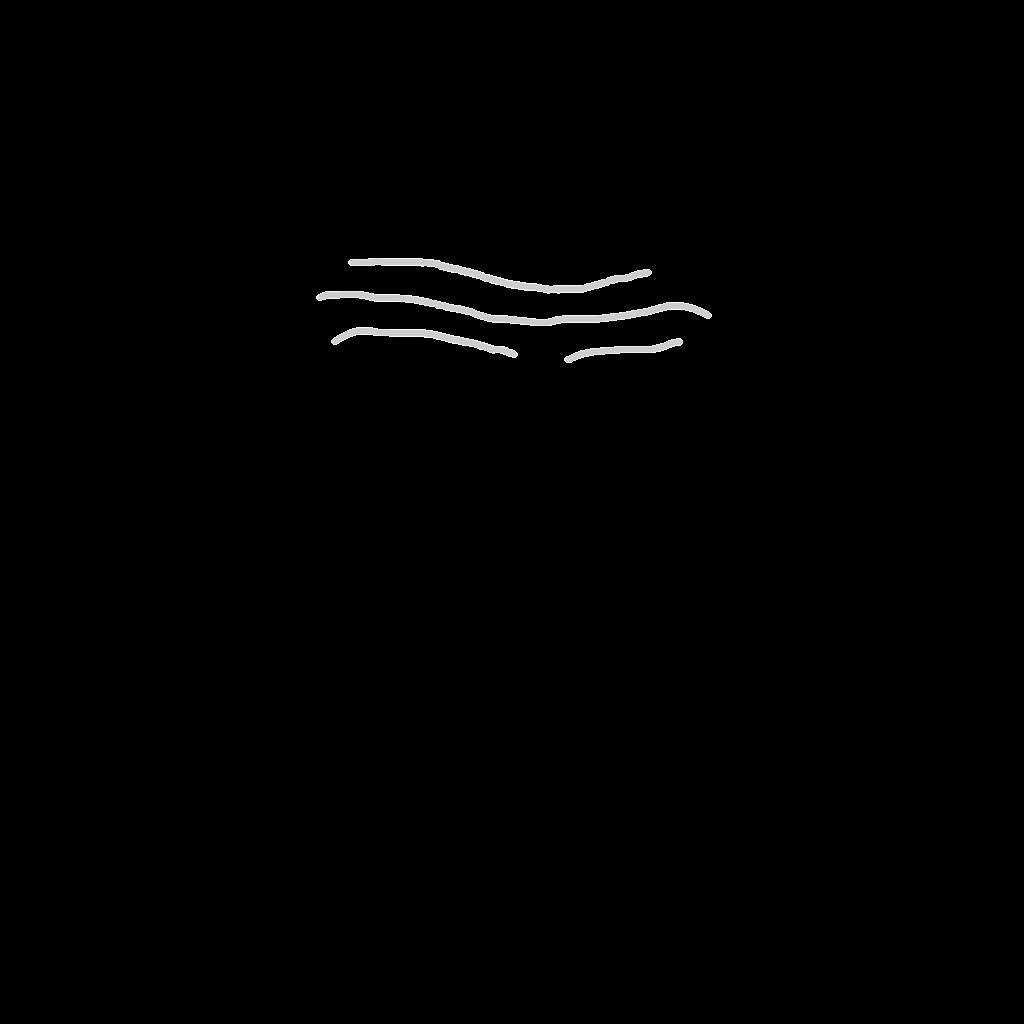} &
     \includegraphics[height=0.2\linewidth]{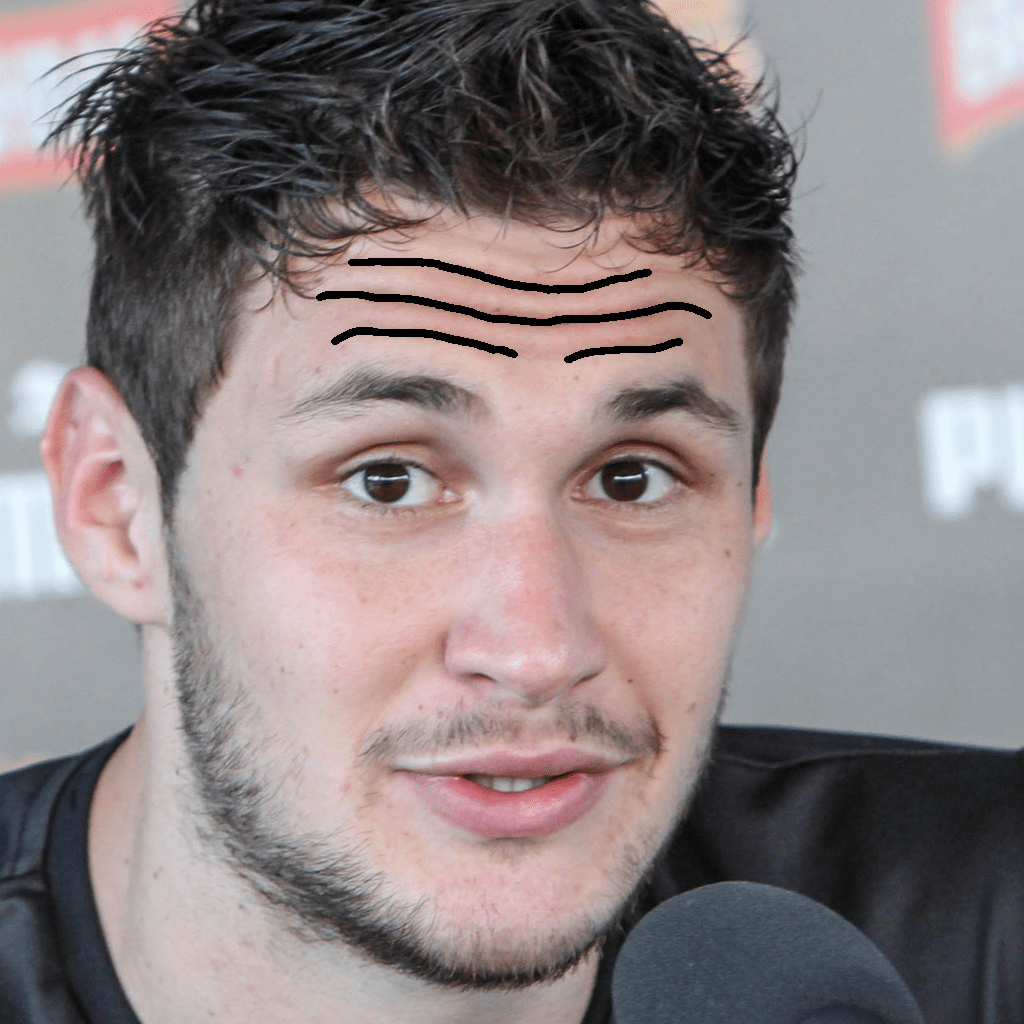} \\
  
        \end{tabular}
        \endgroup
\caption{Samples from FFHQ-Wrinkles dataset.}
\label{fig:FFHQ-Aging}
\end{figure}

We introduce a novel facial dataset to evaluate wrinkle segmentation methods. Based on FFHQ~\cite{karras2019style}, our new dataset, which we dub \textit{FFHQ-Wrinkles}, is composed by 100 images of different subjects paired with corresponding manually annotated wrinkle segmentation masks. In order to decide which samples to annotate, we have used FFHQ-Aging dataset~\cite{or2020lifespan} age metadata to easily identify potential cases. Even though older people are more prone to develop wrinkles, we also added younger people samples with facial expression like the ones showed in Fig.~\ref{fig:FFHQ-Aging} to obtain a more diverse population. We have also ensured to collect samples from different ethnicities and skin tones. All the metrics scores have been computed on the \textit{FFHQ-Wrinkles} dataset.

For training the different methods, we use a proprietary dataset containing 5k samples of RGB frontal face pictures with its corresponding wrinkle annotations. This dataset contains images from subjects diverse in age, gender and ethnicity. Pictures were taken under uncontrolled lightning condition with a native resolution of $ 1024 \times 1024 $. 

In order to assess the different wrinkle removal methods, we adopt the common metrics reported in inpainting literature. We use LPIPS (Learned Perceptual Image Patch Similarity) \cite{zhang2018unreasonable} and FID (Fréchet inception distancee) \cite{heusel2017gans} metrics. However, both LPIPS and FID require a target distribution image to be computed. Since we do not have ground truth images  with wrinkles removed, we propose the following: first, we synthetically create wrinkle-shaped masks that lie on the skin of the subjects, but purposefully avoid overlapping with wrinkles. Then, we evaluate our inpainting method using these masks. Since our model should recover the clear skin, the inpainted result can be directly compared with the original image.

\subsection{Implementation details}

The entire pipeline is implemented using Pytorch~\cite{NEURIPS2019_9015} framework. The segmentation module is based on the implementation from \cite{Yakubovskiy:2019} Unet++ \cite{zhou2019unet++} with a ResNeXt-50 \cite{xie2017aggregated} backbone pre-trained on Imagenet classification task. We train $\Phi_{\rm S}$ for 200 epochs using Adam optimizer \cite{kingma2014adam} with a initial learning rate of 0.001 scaled at 100 epochs by 0.5. Images are downsampled to $ 512 \times 512$, other image transformations such as VerticalFlip, HorizontalFlip, RandomShift from Albumentations~\cite{2018arXiv180906839B} are also applied. We treshold with 0.5 the probability segmentation map in order to set if a given pixel is part of a wrinkle.

The inpainting module is based on the official LAMA implementation \cite{suvorov2022resolution}, the generator is composed by 9 blocks. We use Adam optimizer \cite{kingma2014adam} for both generator and discriminator with a learning rate of 0.0001. We train using $256\times256$ crops with different spatial image transformartion such us Rotation and VerticalFlip. We use a batch size of 16 and train for 300 epochs. In all the experiments, the hyperparameters search is conducted on separated validation dataset. All methods have been trained with our proprietary dataset and following the recommendations suggested by each author. Figures and metrics reported are computed with our new \textit{FFHQ-Wrinkles} dataset.

\begin{figure*}[!ht]
\setlength{\tabcolsep}{1pt}
\begin{tabular}{ccccc}
\includegraphics[height=0.2\linewidth,width=0.2\linewidth]{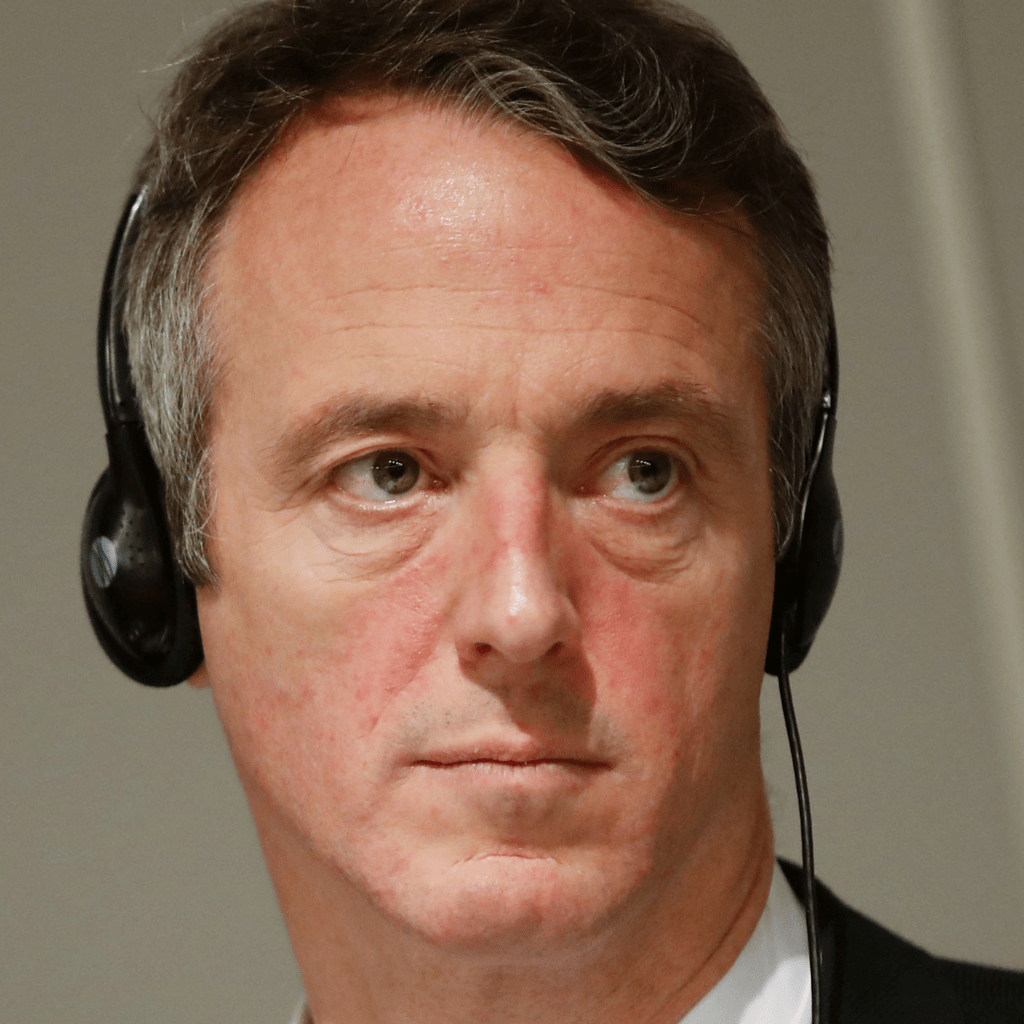} &
\includegraphics[height=0.2\linewidth,width=0.2\linewidth]{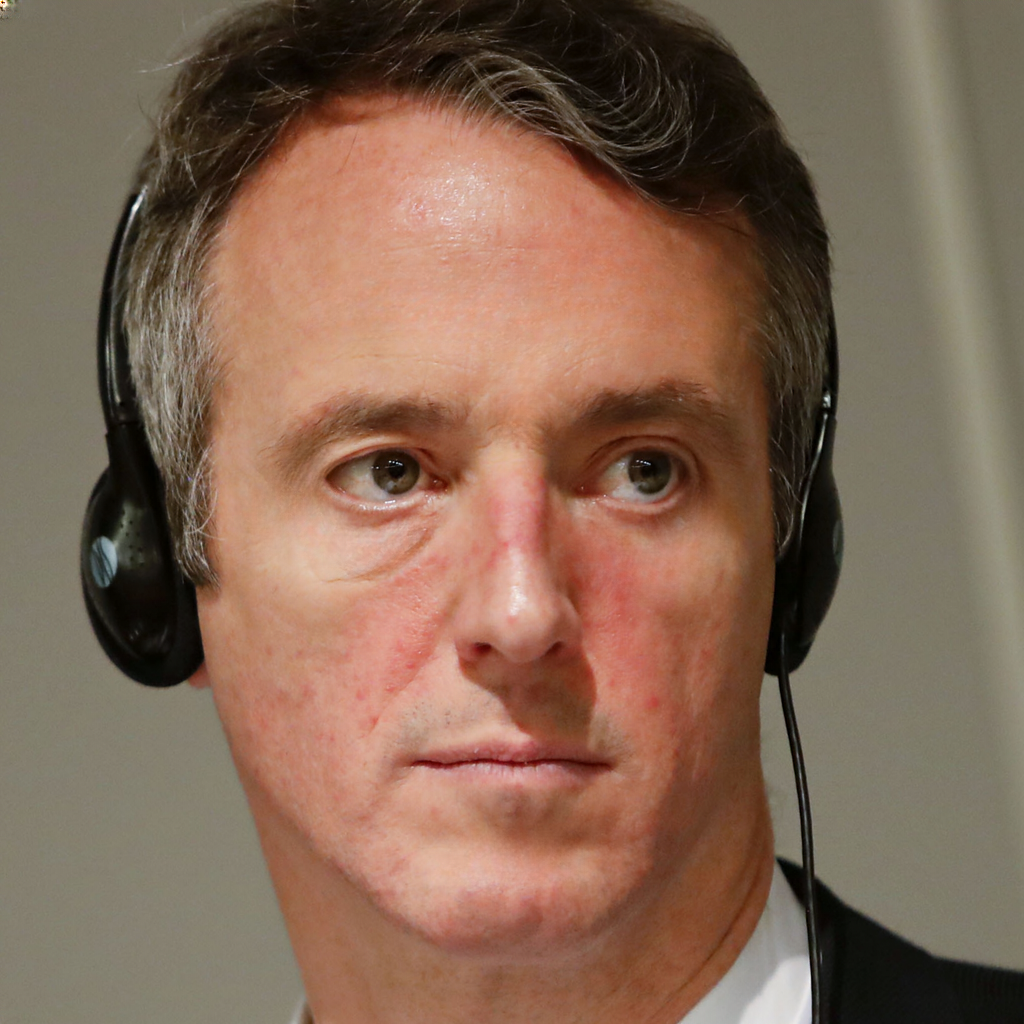} &
\includegraphics[height=0.2\linewidth,width=0.2\linewidth]{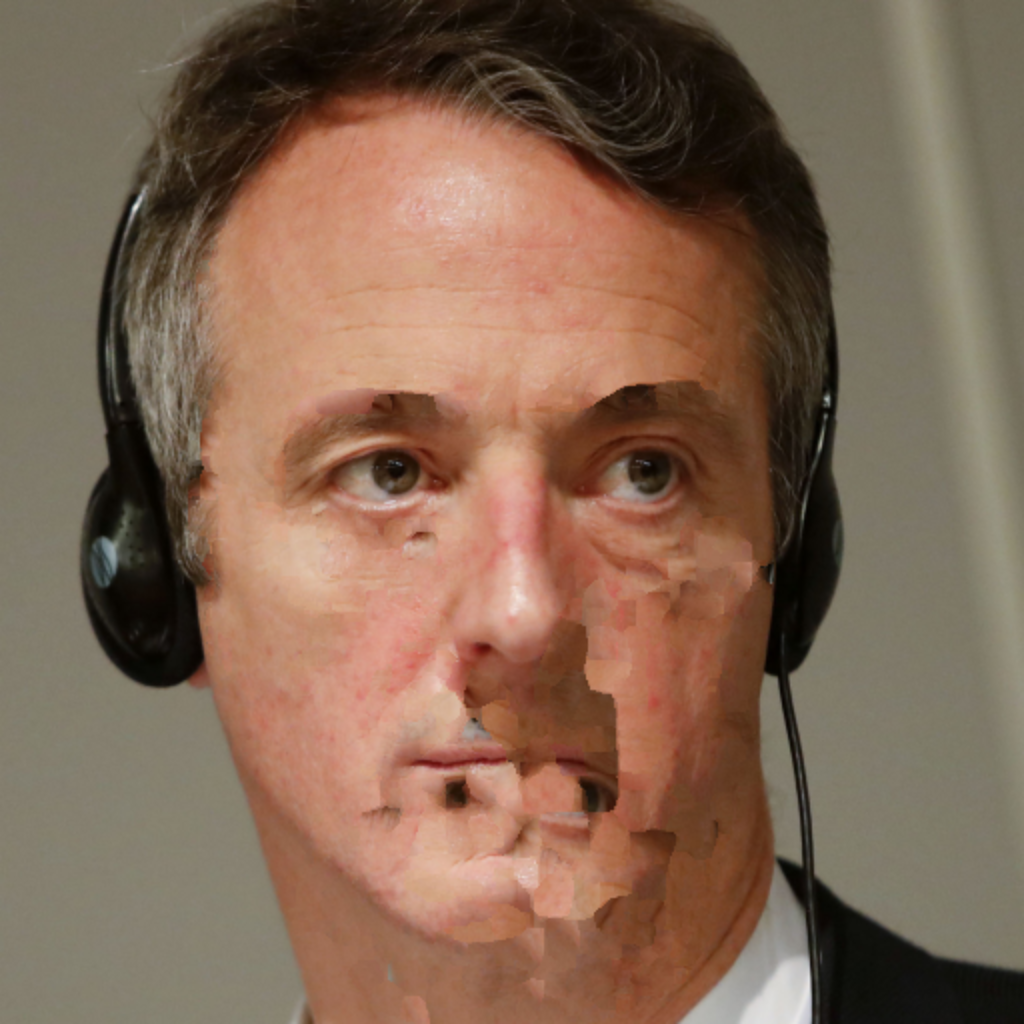} &
\includegraphics[height=0.2\linewidth,width=0.2\linewidth]{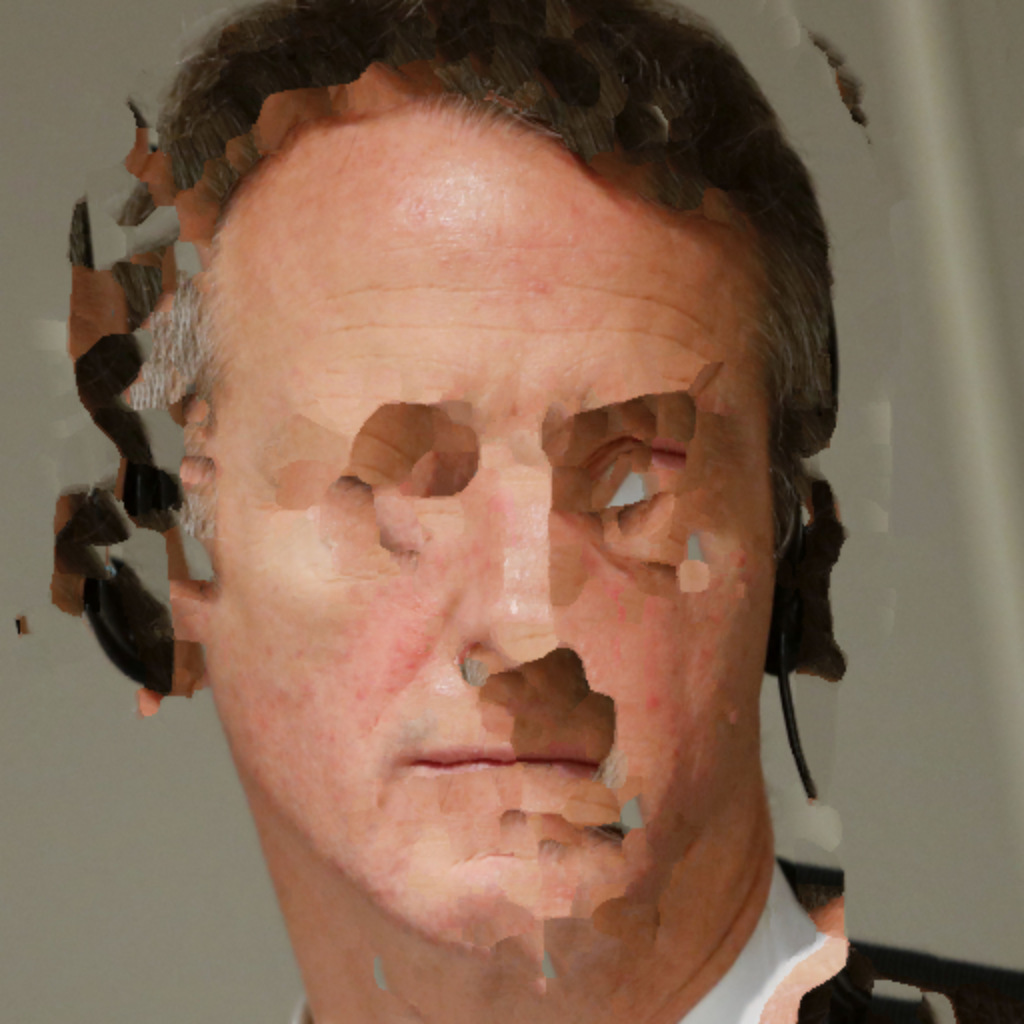} &
\\

&
\includegraphics[height=0.2\linewidth,width=0.2\linewidth]{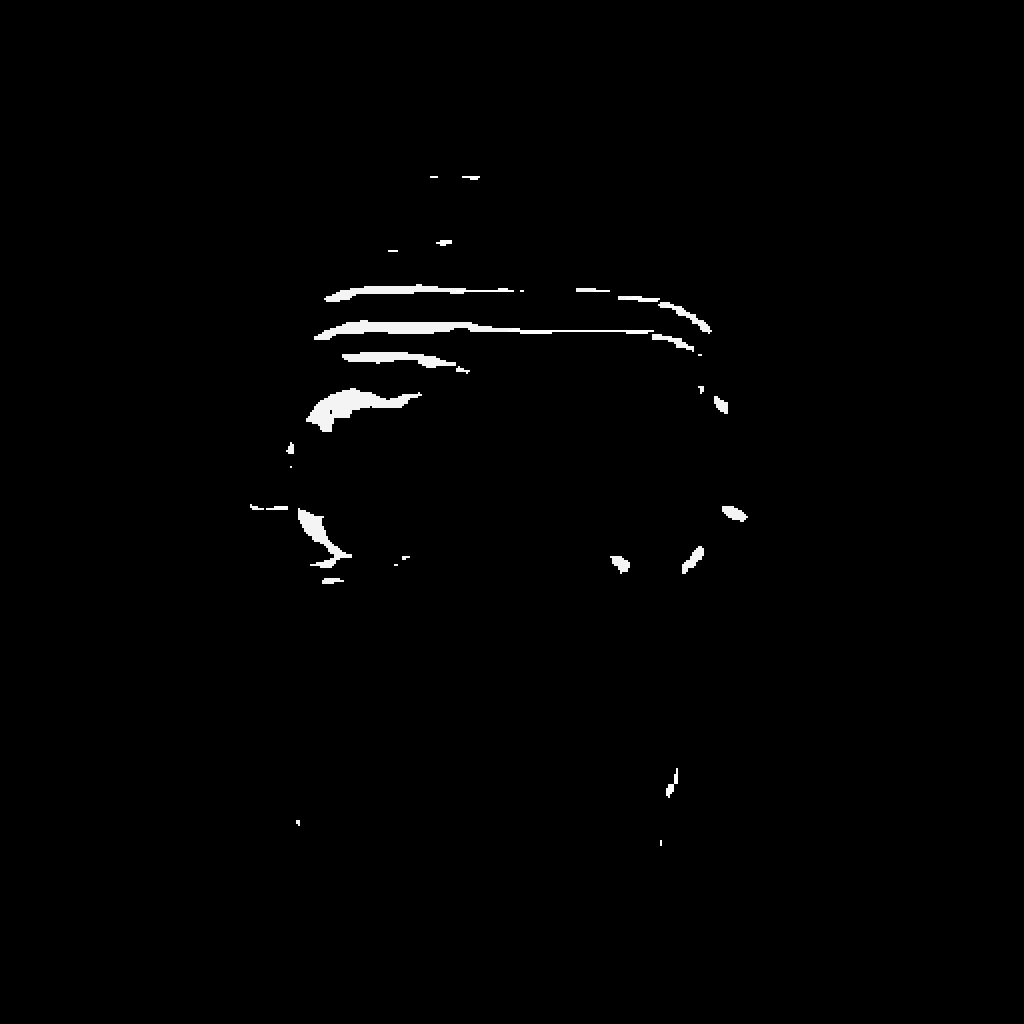} &
\includegraphics[height=0.2\linewidth,width=0.2\linewidth]{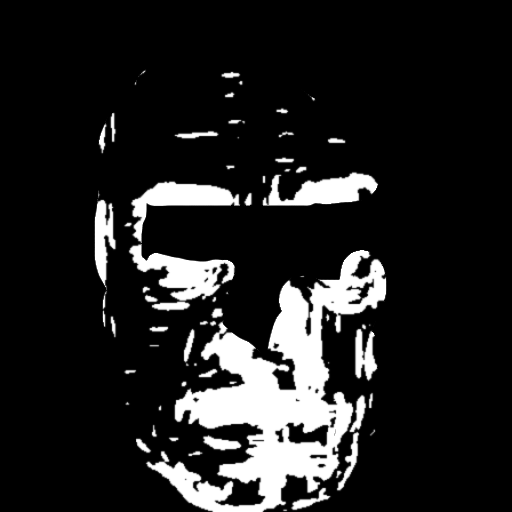} &
\includegraphics[height=0.2\linewidth,width=0.2\linewidth]{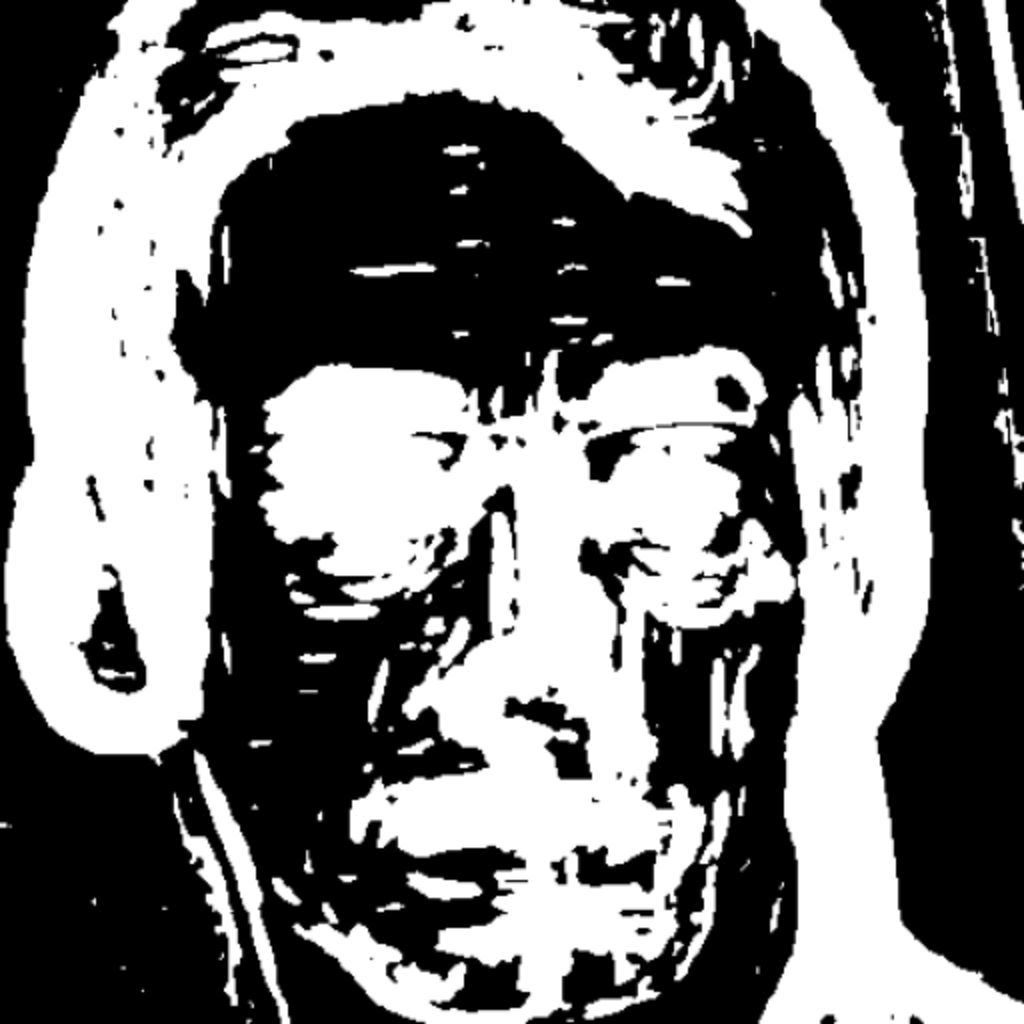} &
\\

\includegraphics[height=0.2\linewidth,width=0.2\linewidth]{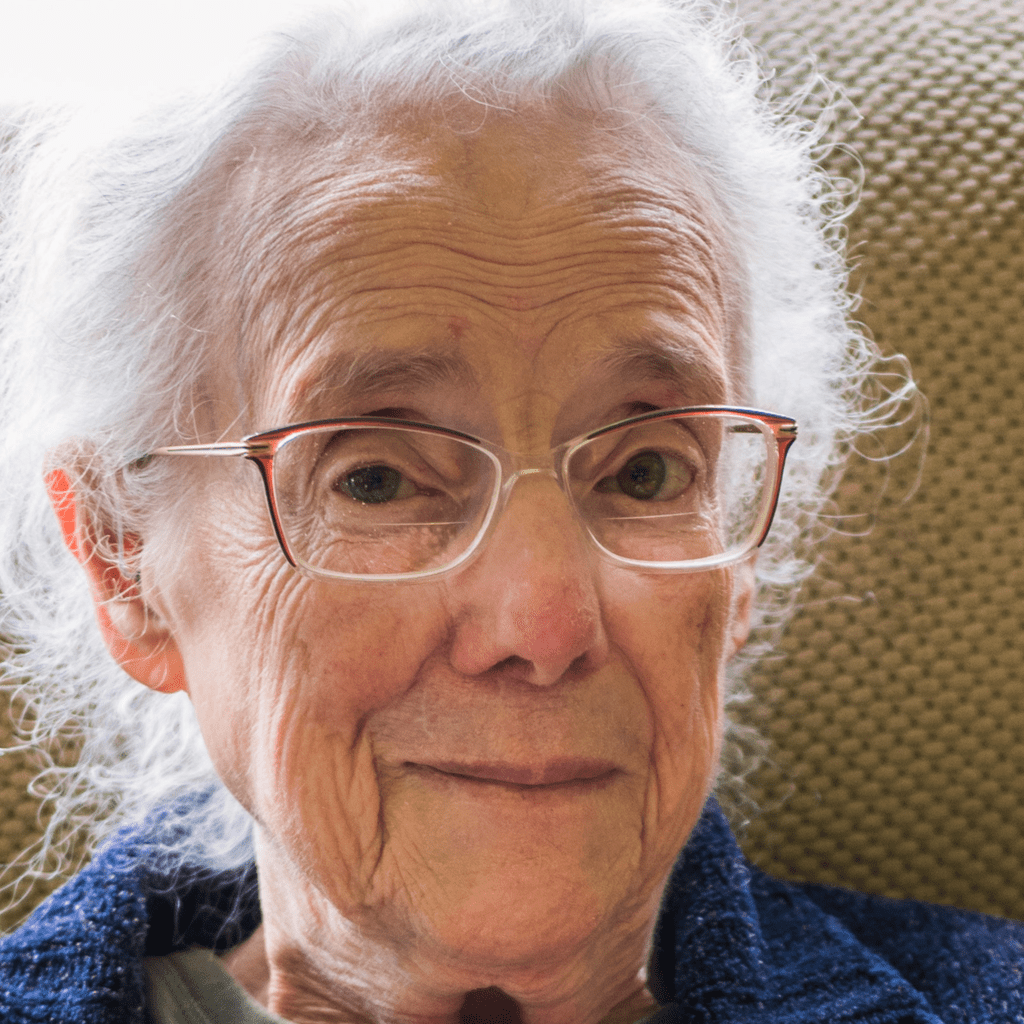} &
\includegraphics[height=0.2\linewidth,width=0.2\linewidth]{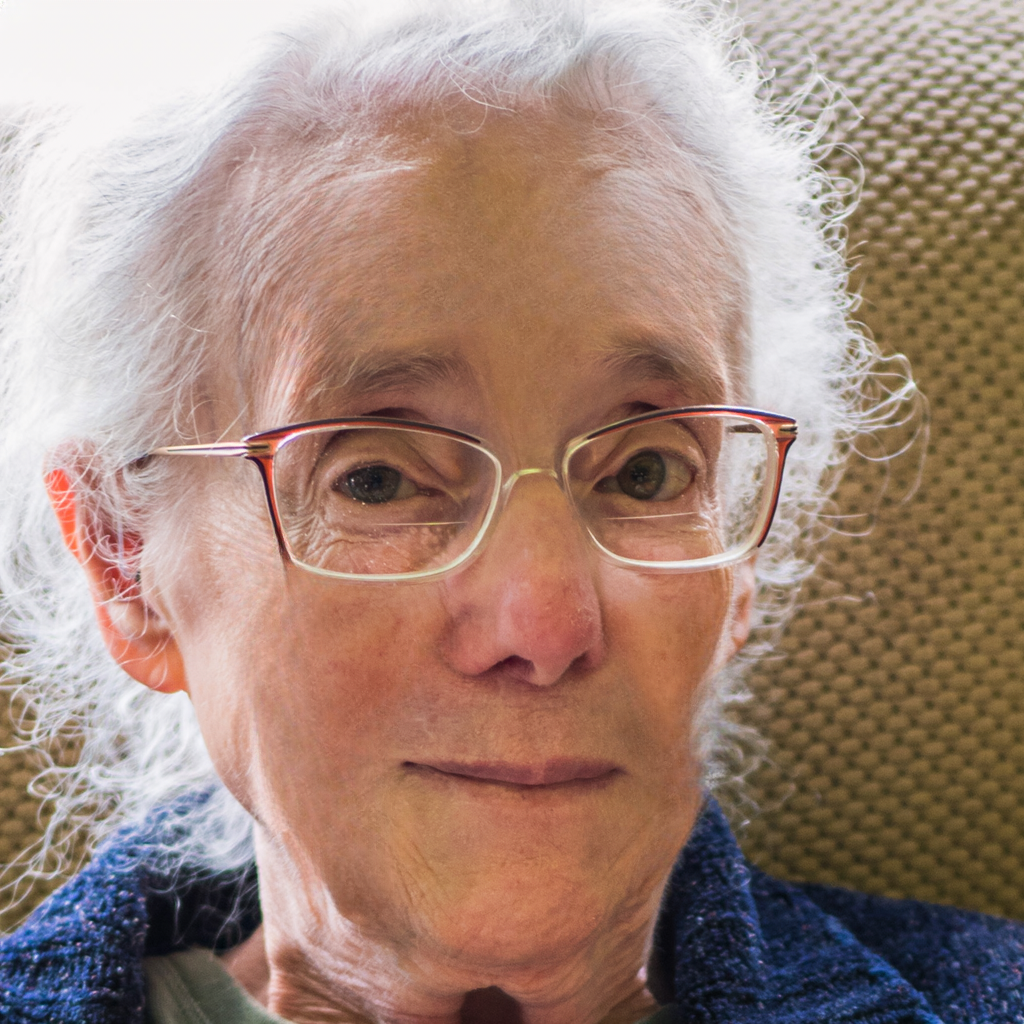} &
\includegraphics[height=0.2\linewidth,width=0.2\linewidth]{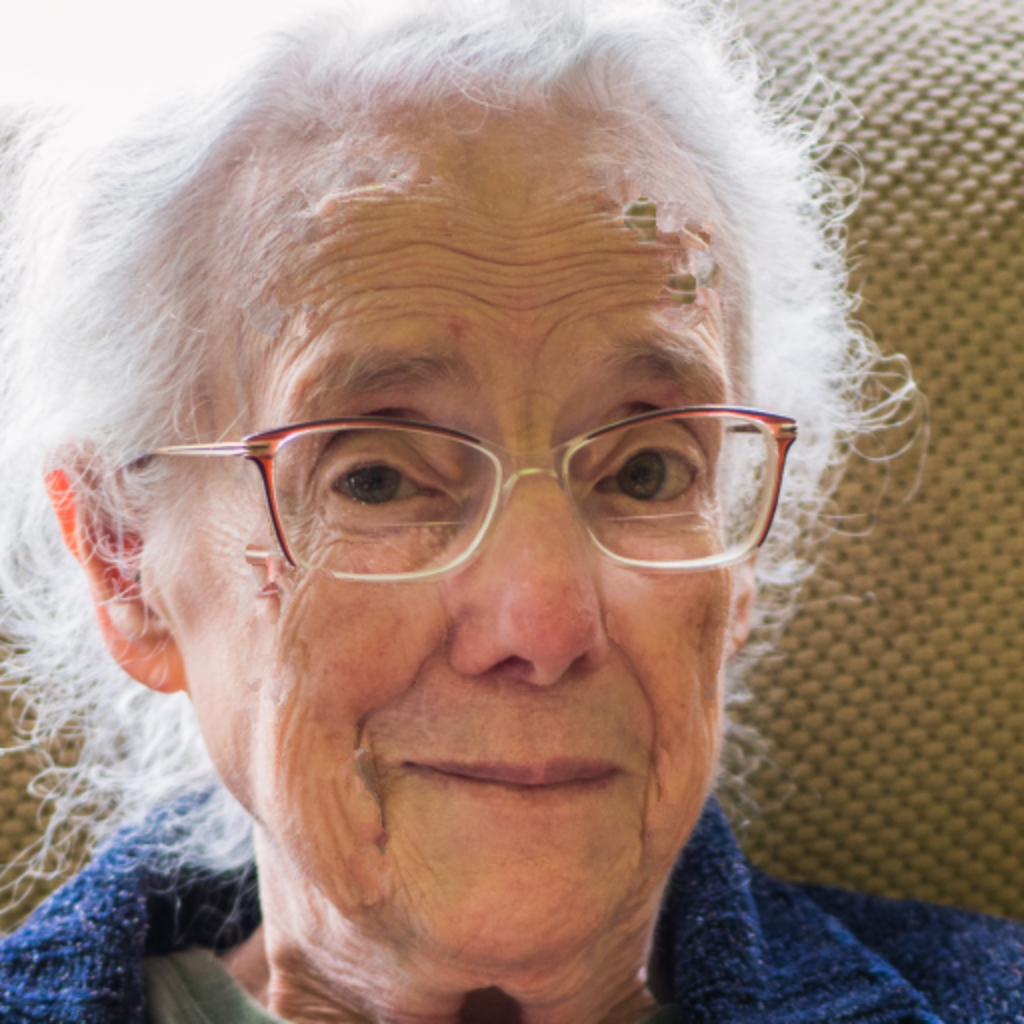} &
\includegraphics[height=0.2\linewidth,width=0.2\linewidth]{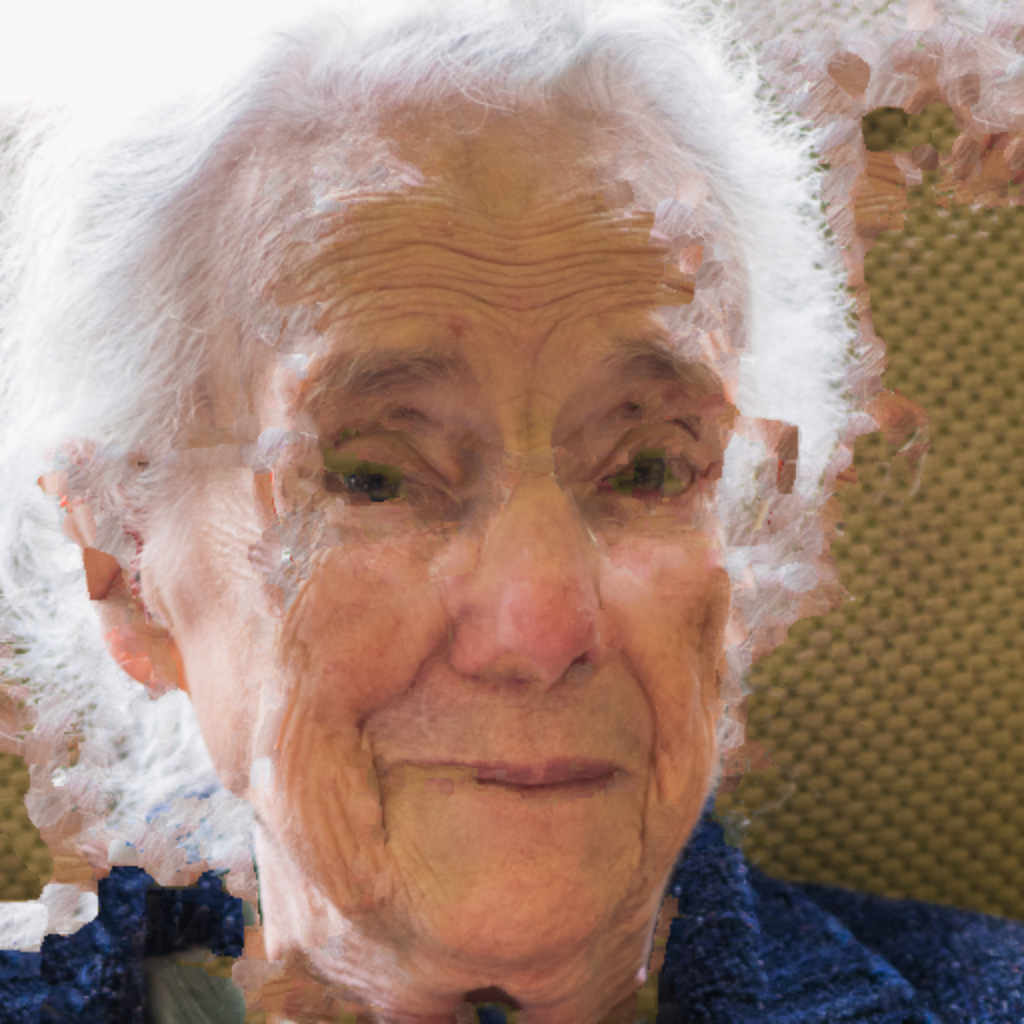} &
\\

&
\includegraphics[height=0.2\linewidth,width=0.2\linewidth]{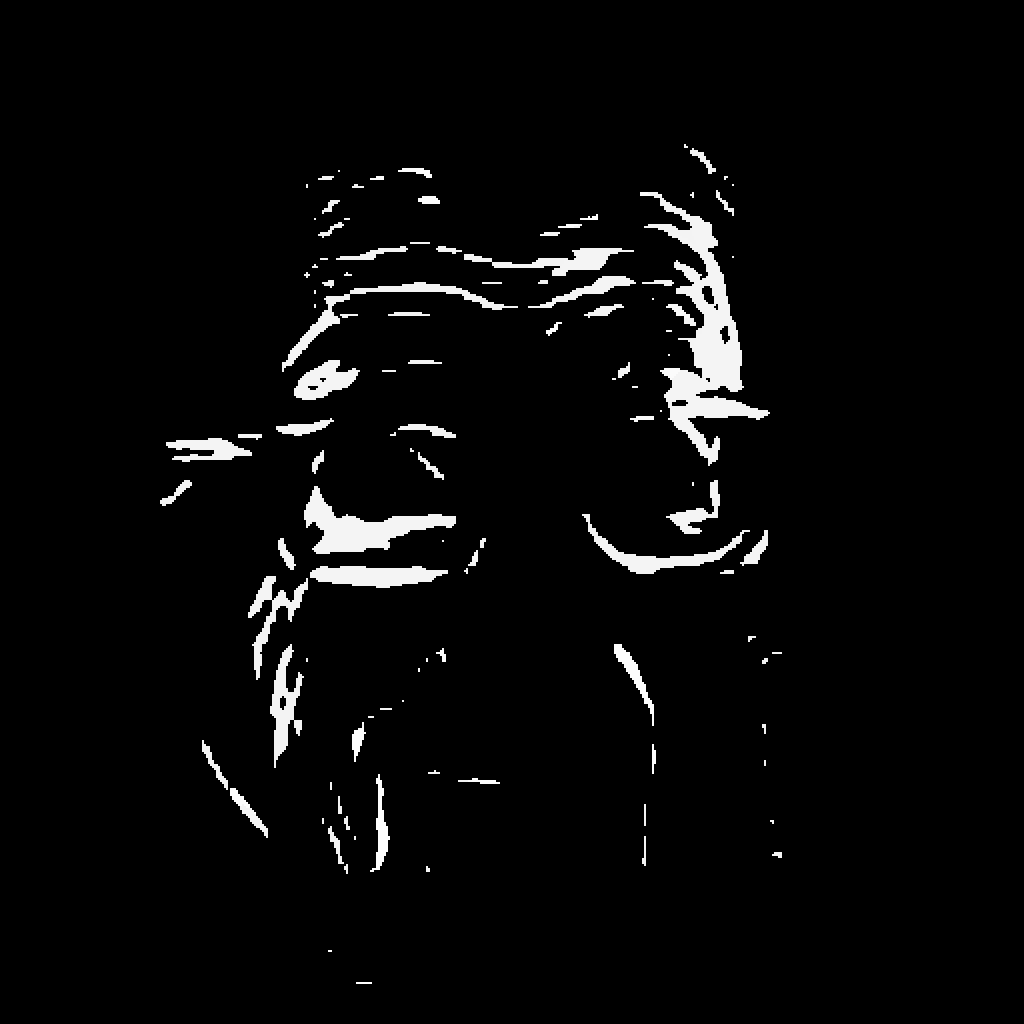} &
\includegraphics[height=0.2\linewidth,width=0.2\linewidth]{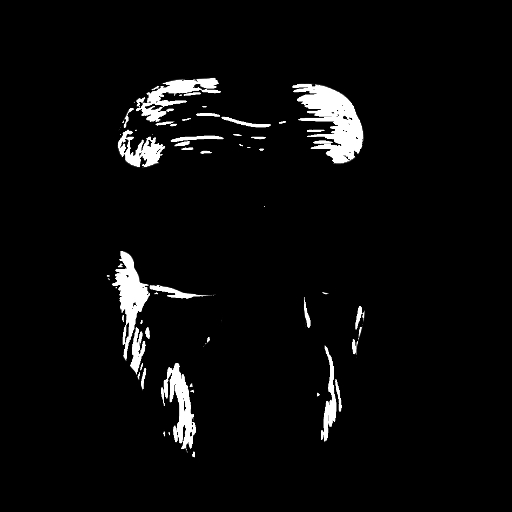} &
\includegraphics[height=0.2\linewidth,width=0.2\linewidth]{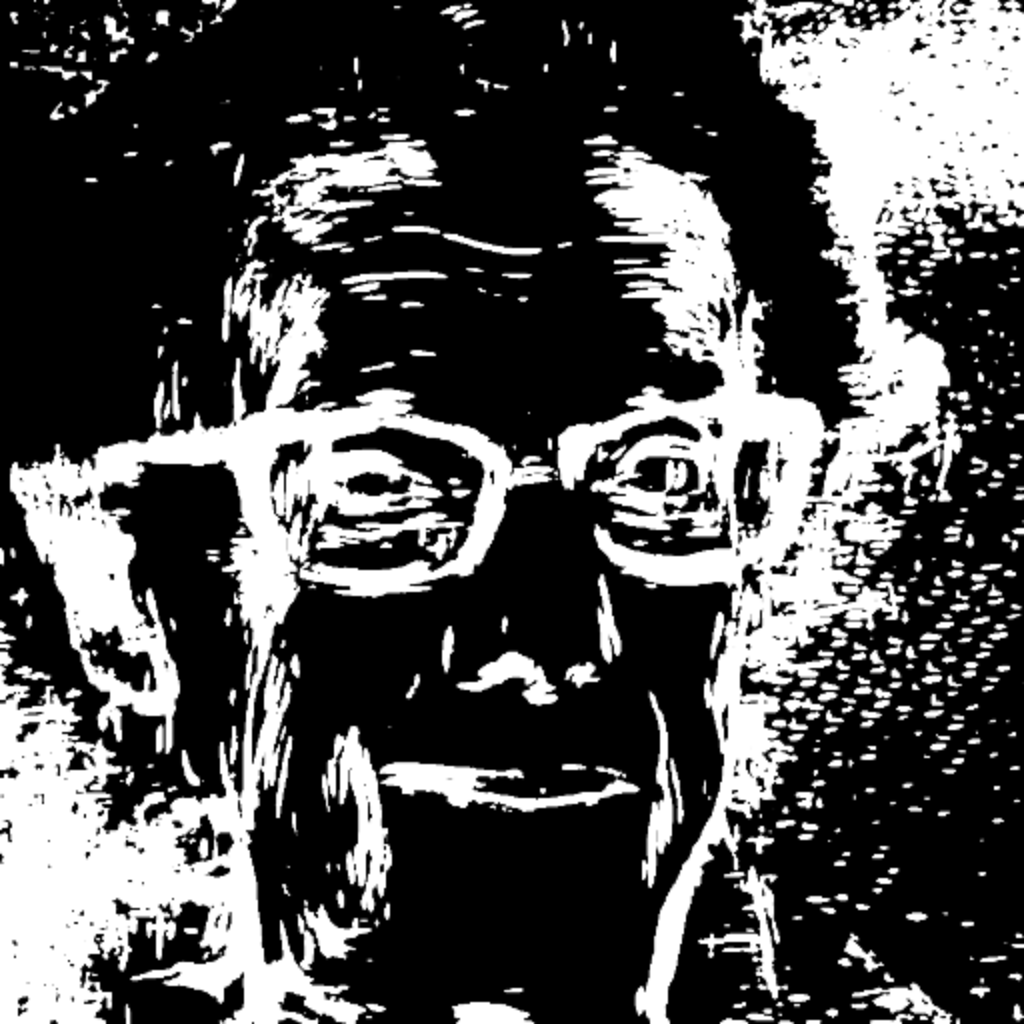} &
\\

\includegraphics[height=0.2\linewidth,width=0.2\linewidth]{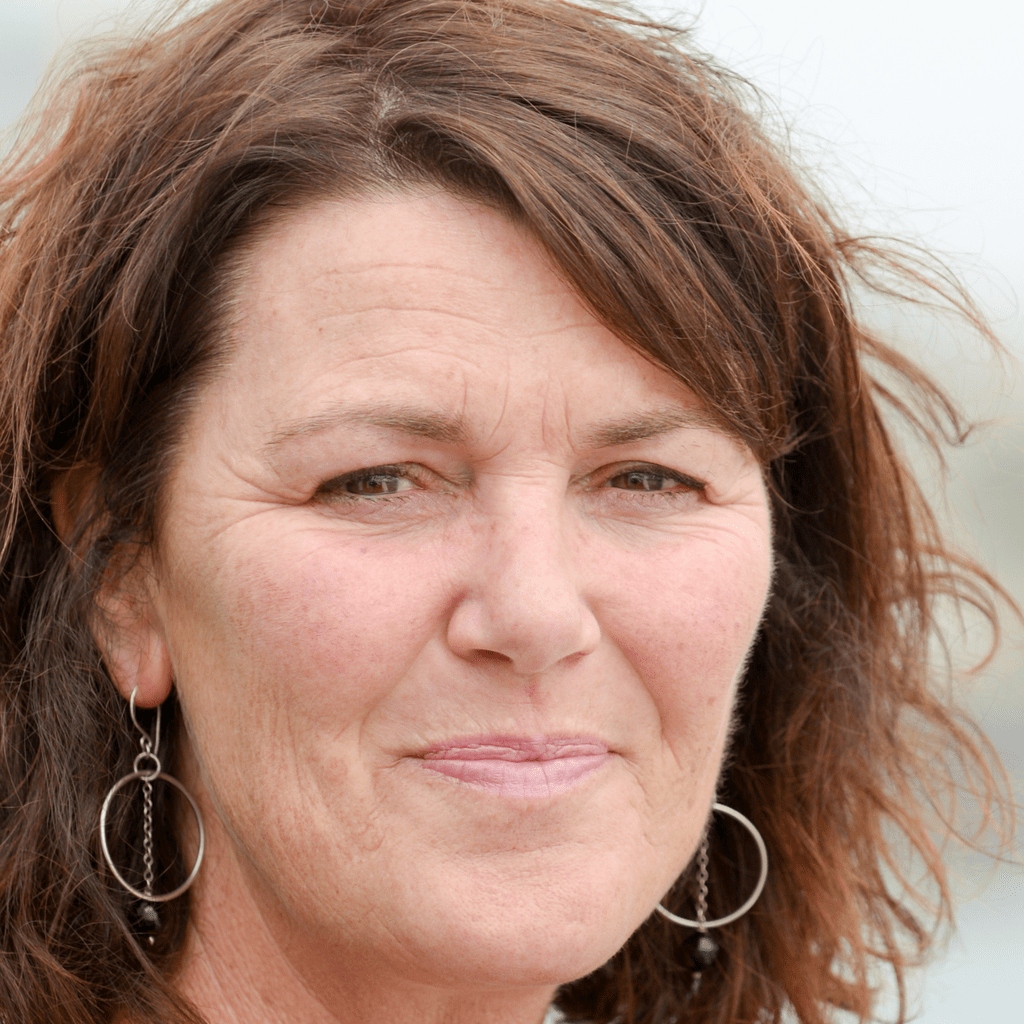} &
\includegraphics[height=0.2\linewidth,width=0.2\linewidth]{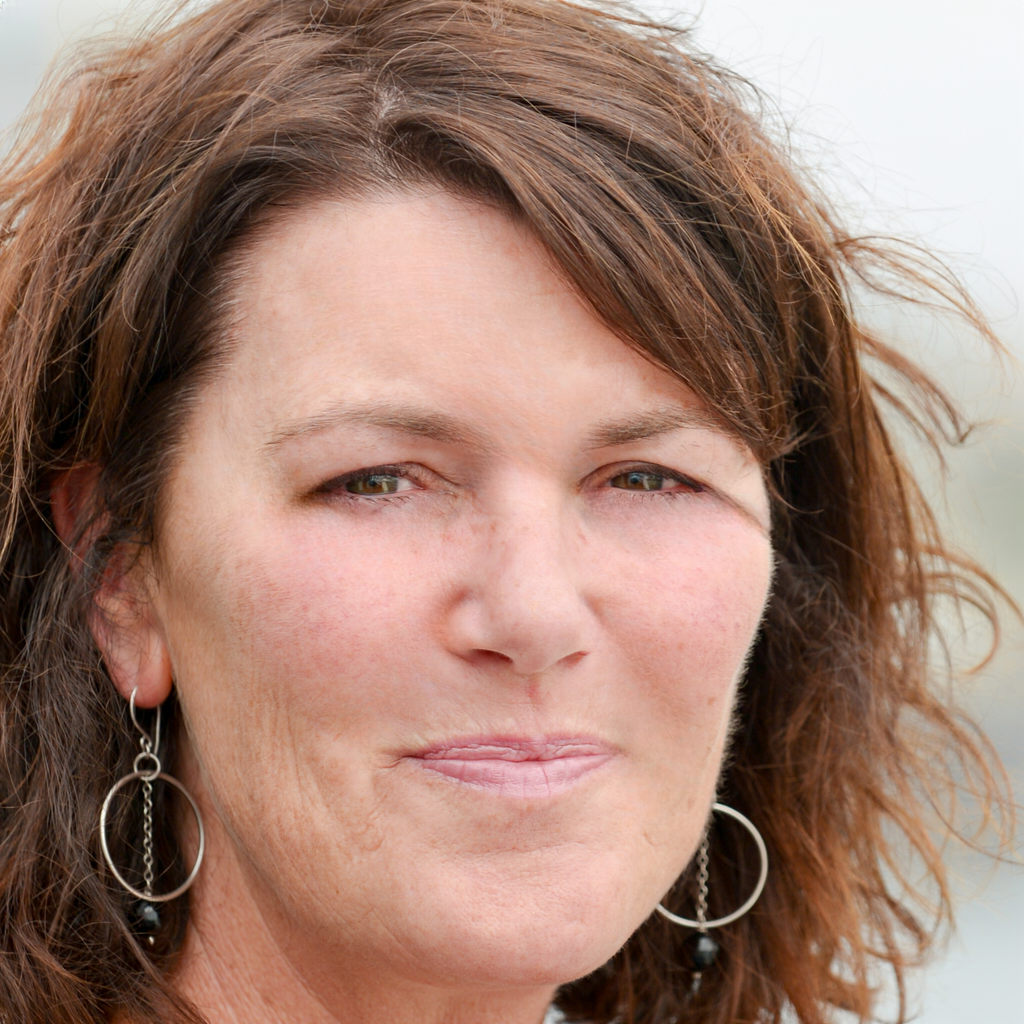} &
\includegraphics[height=0.2\linewidth,width=0.2\linewidth]{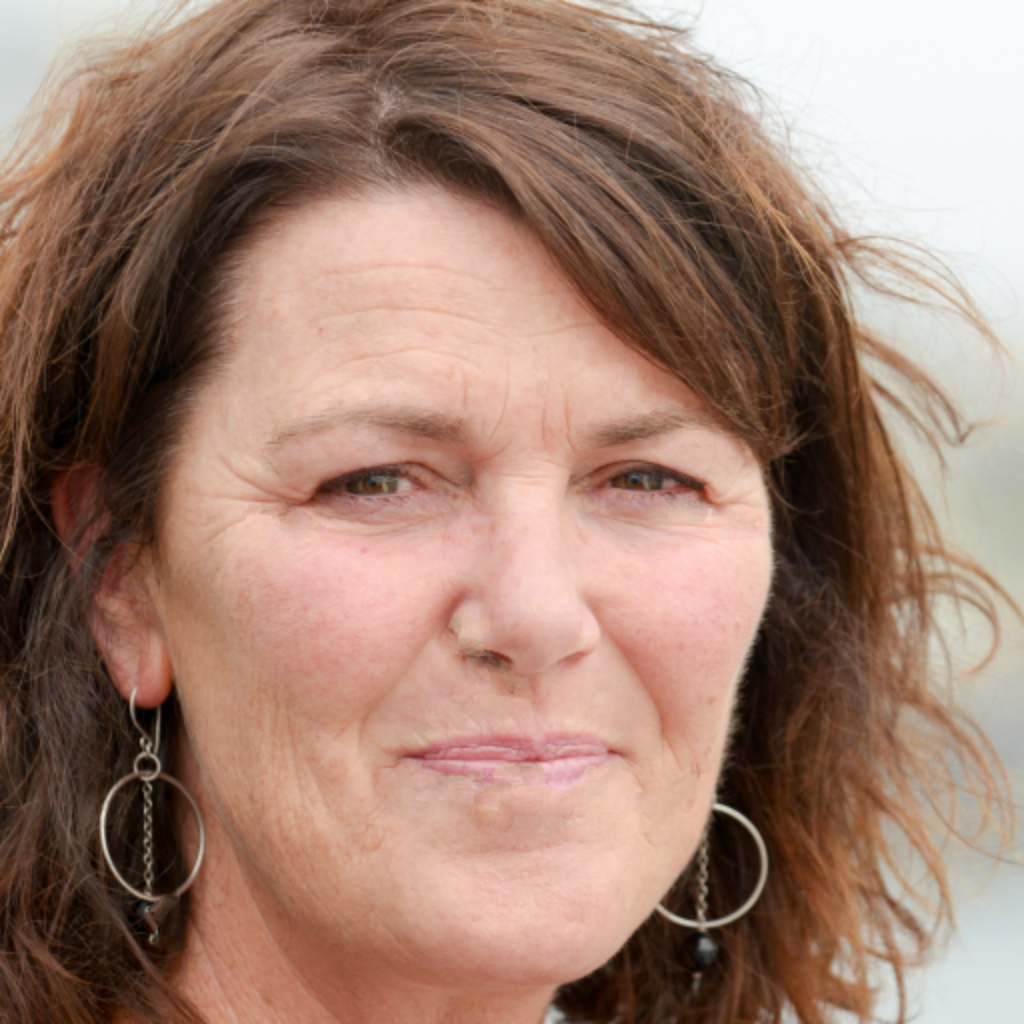} &
\includegraphics[height=0.2\linewidth,width=0.2\linewidth]{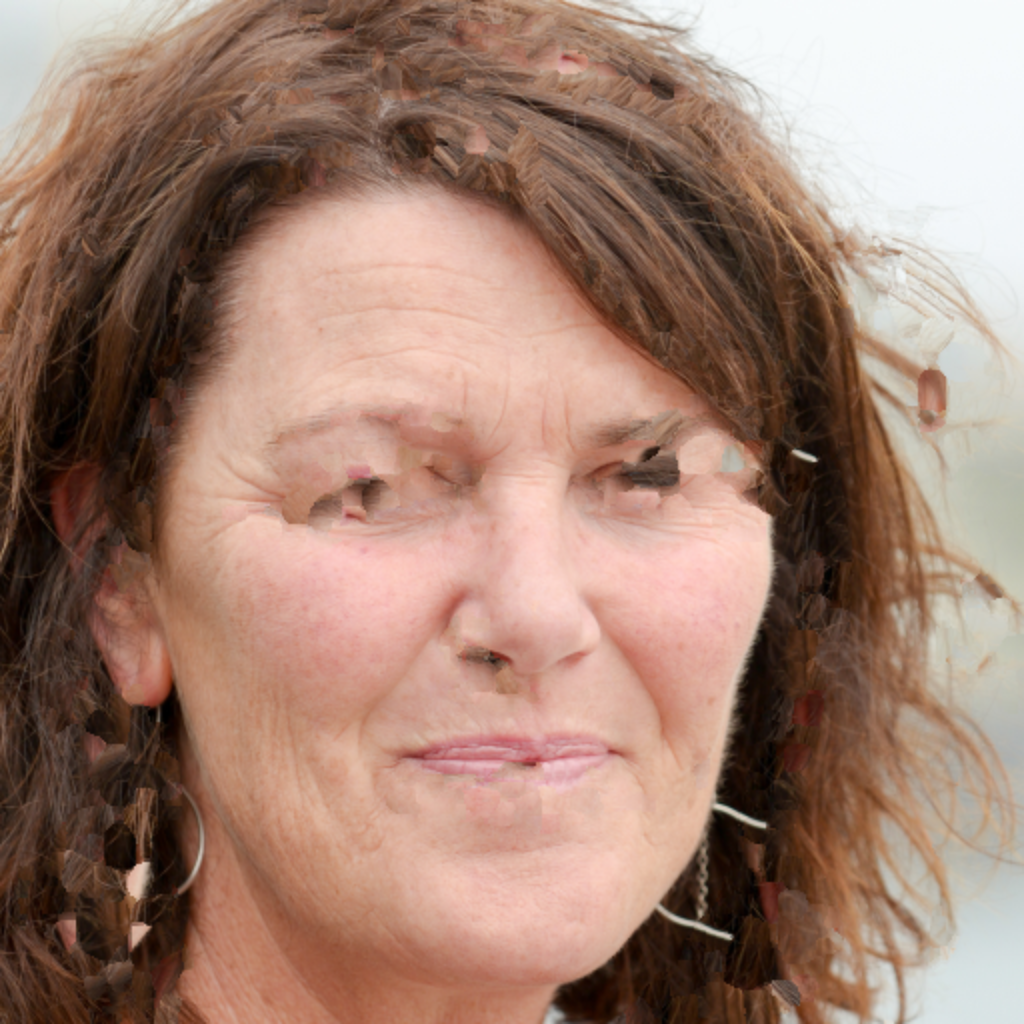} &
\\

&
\includegraphics[height=0.2\linewidth,width=0.2\linewidth]{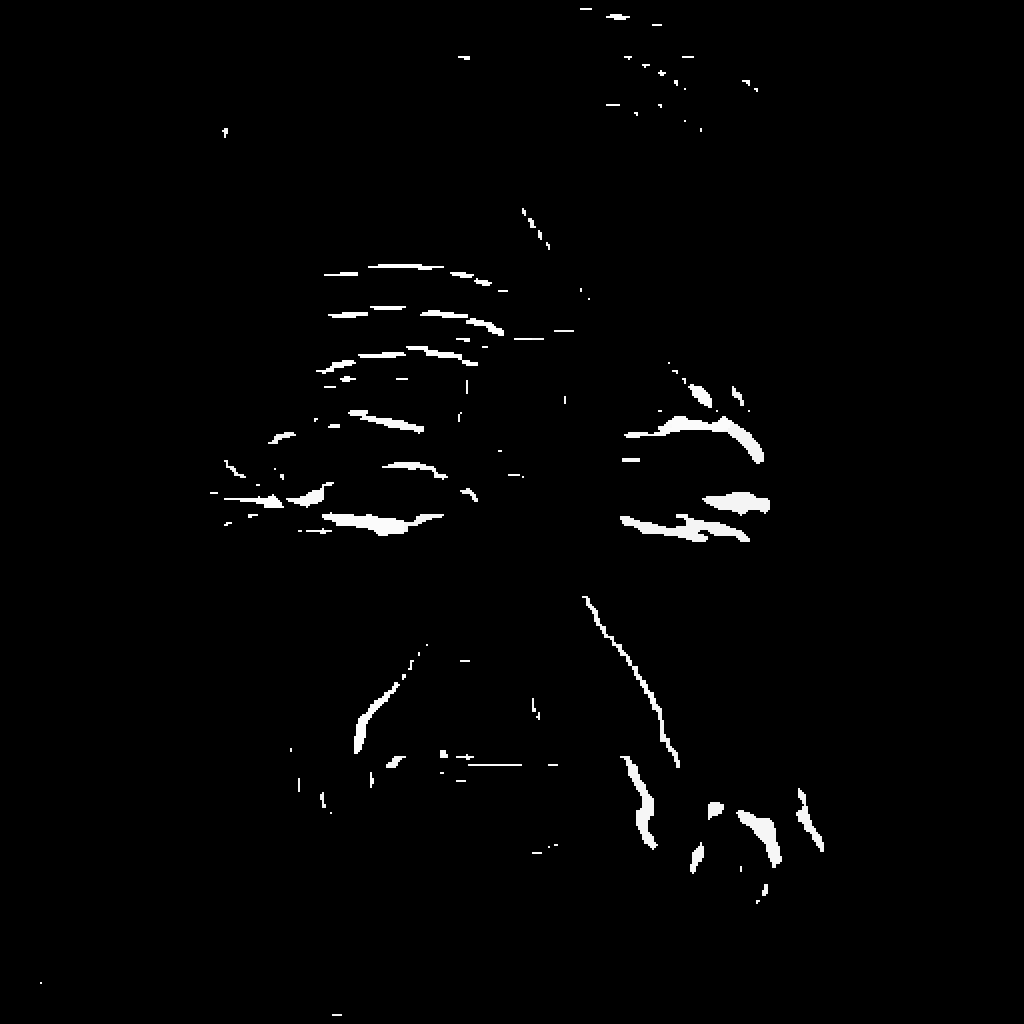} &
\includegraphics[height=0.2\linewidth,width=0.2\linewidth]{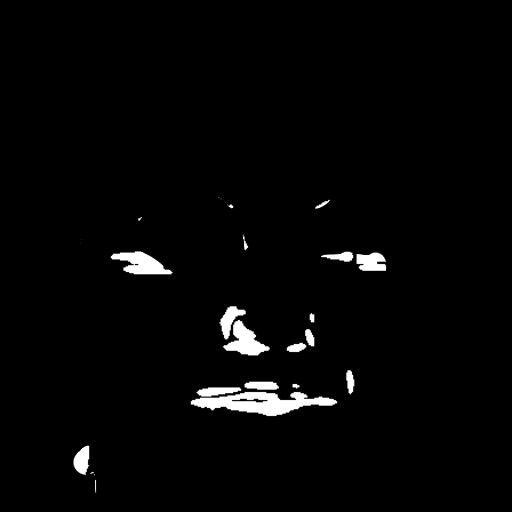} &
\includegraphics[height=0.2\linewidth,width=0.2\linewidth]{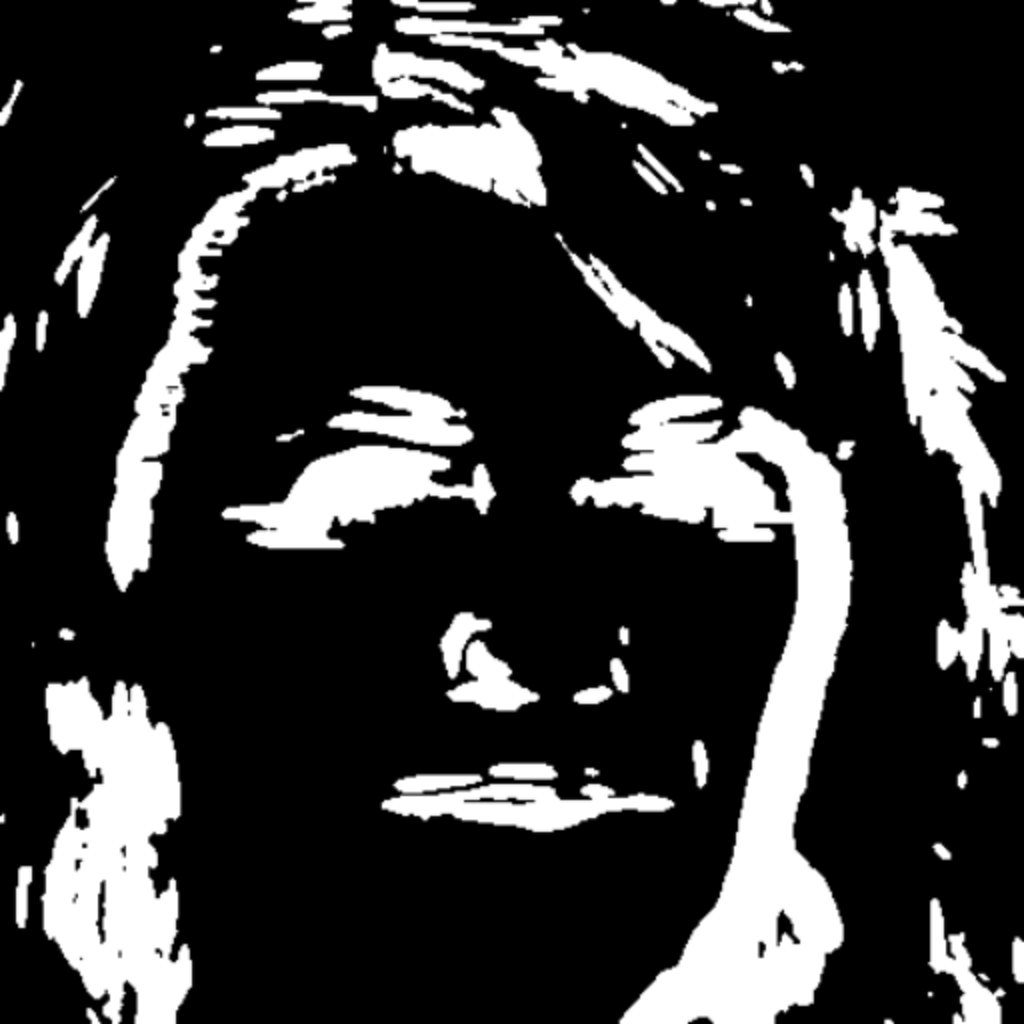} &
\\

Image & Ours & \makecell{Batool et al \cite{batool2014detection}  \\ with manual area \\ selection} & \makecell{Batool et al \cite{batool2014detection} \\ automatic}
\end{tabular}
\caption{Qualitative comparison to the state-of-the-art methods on wrinkle cleaning on FFHQ-Wrinkles dataset. Our methods outperforms current state-of-the-art algorithms and is robust to external elements of the face such us hair, clothes or earrings.}
\label{fig:inpainting-comparative sota}
\end{figure*}

\subsection{Comparisons to the baselines}
We compare the proposed approach on wrinkle cleaning with state-of-the-art methods in Fig.~\ref{fig:segmenation-results}. Batool et al \cite{batool2014detection}  method contrary to ours needs human intervention to obtain correct wrinkle segmentation. This intervention is done by selecting the potential facial regions containing wrinkles in order to reduce the search space. Without human intervention it tends to segment as wrinkles high gradient sections such us hair or clothes. In order to  provide a fair comparison and favour their results, we compare our method with two variants of \cite{batool2014detection}, which are displayed in Fig.~\ref{fig:segmenation-results}. The first variant consists in using the segmentation map obtained directly from their detection module (with no user intervention). The corresponding 
results with the segmentation map obtained by \cite{batool2014detection} are shown in the right-most column. For each experiment, the inpainting result is shown in the first row and the segmentation result in the second row. The second variant consists in manually filtering that segmentation mask obtained automatically by \cite{batool2014detection} by applying a mask marking the region of valid wrinkles (i.e., avoiding hair, glasses, \dots). As we do not have access to the authors' inpainting module, we used a patch based inpainting approach~\cite{ipol.2017.189}. similar to the one used in~\cite{batool2014detection} which is based on the texture synthesis method~\cite{efros2001image}.

Our model largely outperforms method \cite{batool2014detection} and is robust to external elements of the face such us hair, clothes or earrings. Also is able to fill the wrinkle regions with a more natural and smooth skin, leading to skin patches indistinguishable to the human eye.

\begin{table}[h!]
  \begin{center}
\begin{tabular}{|l|c|c|c|}
\hline
 Configuration &  IoU $\uparrow$ & LPIPS $\downarrow$ & FID $\downarrow$\\ 
\hline\hline
 {Ours} 
  & 0.561 & 0.231  & 7.24 
 \\ 
  {Batool et al ~\cite{batool2014detection}}
  & 0.192 & 3.43 & 14.11\\ 
  \hline
\end{tabular}
\end{center}
  \caption{Quantitative comparison with previous state-of-the-art wrinkle cleaning method on FFHQ-Wrinkles dataset.}
  \label{tab:segmentation-res}
\end{table}

\section{Ablations}
\input{accv2022/sections/ablations}
    
    
\section{Conclusions}
\label{sec:conclusions}
We have presented a pipeline for photorealistic wrinkle removal. This pipeline is based on state-of-the-art methods for segmentation and inpainting, achieving unprecedented results. The proposed two-stage pipeline pushes the state of the art in wrinkle cleaning however, struggles with non-frontal images. Our experiments on FFHQ-Wrinkles demonstrate the effectiveness of our new proposed loss and how it helps to fill regions with plausible skin. We also provide the first public dataset that for wrinkle segmentation in orde to ease future work on this topic and provide a baseline.  We believe that solving wrinkle detection and wrinkle cleaning independently is a key feature that adds interpretability to the system. However, exploring ways to jointly fuse this pipeline and train jointly is an exciting and interesting future work.

\bibliographystyle{splncs}
\bibliography{accv2022/egbib}

\end{document}